\documentclass[preprint,12pt]{elsarticle}

\usepackage{natbib}
\usepackage{graphicx}
\usepackage{booktabs}
\usepackage{array}
\usepackage{multirow}
\usepackage{amsmath}
\usepackage{amssymb}
\usepackage{microtype}
\usepackage{url}
\usepackage{hyperref}
\usepackage{xcolor}
\usepackage{amsthm}
\usepackage{tikz}
\usepackage{longtable}
\usepackage[section]{placeins}
\usetikzlibrary{shapes.geometric,arrows.meta,positioning,calc,fit,backgrounds}

\newtheorem{remark}{Remark}

\newcommand{\vtot}{V_{\text{total}}}
\newcommand{\cone}{V_1}
\newcommand{\ctwo}{V_2}
\newcommand{\cthree}{V_3}
\newcommand{\ka}{Krippendorff's~$\alpha$}
\newcommand{\fk}{Fleiss'~$\kappa$}
\newcommand{\ck}{Cohen's~$\kappa$}
\newcommand{\icc}{\text{ICC}(A,1)}

\journal{Knowledge-Based Systems}

\begin{document}

\begin{frontmatter}

\title{Measurement Without Validity: The Compounding Reliability Problem in Agentic AI Evaluation}

\author[addr1]{William Caban\corref{cor1}}
\ead{wcabanba@redhat.com}

\cortext[cor1]{Corresponding author. ORCID: 0009-0008-1211-7050}

\address[addr1]{Red Hat, Inc., Raleigh, North Carolina, USA;
  Alma Mater Europaea University, Vienna, Austria}

% -----------------------------------------------------------------------
\begin{abstract}
Agentic AI evaluation pipelines produce benchmark scores that justify deployment
decisions, safety certifications, and regulatory compliance claims. No formal
framework has characterized how validity degrades across these stages. We
present a knowledge-based three-layer compounding validity model,
$V_{\text{total}} \leq V_1 \times V_2 \times V_3$, that captures multiplicative
degradation across task generation ($V_1$), human-simulator calibration ($V_2$),
and automated judgment ($V_3$). Applying it to five widely-used agentic
benchmark configurations, we find $V_1$ and $V_2$ independently measurable and
both degraded, but $V_3$ not currently quantifiable for any: three of the four
benchmarks score with a single deterministic rule, so no second rater exists and
reliability is undefined, and the fourth's partial LLM grader has no published
grader-vs-human measurement.

A structured survey of 55 published agentic evaluation papers finds that 26 of
them report no chance-corrected inter-rater statistic of any kind, a
determination four independent raters recover at Krippendorff's
$\alpha = 0.834$. On the broader
judgment of whether a reported metric fits its rating design, approximately
80\% of this purposive sample use structurally mismatched, incomplete, or absent
inter-rater reliability (IRR) metrics (four-rater range 60.7\%--84.4\%). Both
are consistent with systematic $V_3$ collapse. We further document $V_1$
failures (task validity flaws in 7 of 10 benchmarks) and $V_2$ miscalibration
(up to 9 percentage points inter-simulator variance, with systematic demographic
disparities for non-Standard American English speakers).

We derive eight prescriptions grounded in psychometric science and
domain-stratified reliability thresholds (intraclass correlation $\geq$0.70;
$\alpha \geq 0.67$/$0.70$/$0.80$ by consequence level) that practitioners can
apply immediately.
\end{abstract}


\end{frontmatter}

% -----------------------------------------------------------------------
\section{Introduction}
\label{sec:intro}

Deployment decisions, safety certifications, and regulatory compliance claims for
agentic AI systems are all downstream of a number that appears deceptively clean: a
benchmark score. That score is the product of a pipeline (task generation, world
simulation, rater judgment), and each stage of that pipeline is a potential source of
systematic, non-random measurement error.

The measurement problem in agentic evaluation is not simply that individual
components are imperfect. It is that the imperfections compound. A pipeline that
loses 30\% of valid signal at task generation, 20\% at simulation, and 25\% at the
judgment layer does not produce evaluation that is 75\% valid. Under the
independence bound, it produces evaluation that retains less than 42\% of valid
signal against the intended construct. This bound is itself optimistic for
same-provider pipelines, where shared systematic biases not captured in the
per-layer estimates cause them to overstate true validity
(\ref{app:compounding}, \S{}B.4).

This paper makes three contributions. First, we document empirically that agentic
evaluation fails three distinct psychometric validity tests simultaneously, and that
the empirical evidence for each failure is published, peer-reviewed, and specific.
Second, we introduce a heuristic three-layer compounding validity model that
characterizes how failures multiply rather than add, a relationship the field has
not yet quantified, and which we present as a conceptual bound rather than a proved
theorem. Third, we
derive eight prescriptions from the psychometric literature that practitioners and
benchmark authors can apply immediately.

We are not arguing that agentic evaluation is worthless. We are arguing that it is
less valid than current practices acknowledge, that the mechanisms of invalidity are
known, and that the standards for reporting and interpreting evaluation results should
be substantially higher.

% -----------------------------------------------------------------------
\section{Background: The Psychometric Standard}
\label{sec:background}

This section introduces the two psychometric constructs that underpin the analysis:
construct validity, which defines what a measurement system is supposed to capture,
and inter-rater reliability, which quantifies the consistency of the scoring process.
Both are necessary conditions for a valid benchmark score.

\subsection{Construct Validity as the Organizing Concept}
\label{sec:construct-validity}

The foundational criterion for any measurement system is construct validity: whether
the measurement instrument actually captures the phenomenon it claims to measure
\citep{cronbach1955}. The measurement science tradition identifies three evaluable
validity types that together constitute evidence for or against construct validity:

\begin{itemize}
  \item \textbf{Content validity}: does the evaluation cover the full scope of the
    construct, or only an easily measurable subset? Does a benchmark described as
    ``measuring agentic tool use'' sample the full range of tool-use behaviors, or
    does it measure only the subset most amenable to automated scoring (e.g., API call
    success rate)?
  \item \textbf{Criterion validity} (concurrent and predictive): does the evaluation
    score correlate with an external criterion of the target construct? Predictive
    criterion validity specifically asks whether the score predicts real-world
    deployment performance.
  \item \textbf{Construct validity evidence} (convergent and discriminant): does the
    evaluation correlate with conceptually related measures (convergent evidence) and
    fail to correlate with conceptually unrelated ones (discriminant evidence)?
\end{itemize}

Despite its centrality to measurement science, construct validity has received
limited attention in AI evaluation, even as benchmark creation involves numerous
high-stakes design decisions including target user selection,
construct-to-task operationalization, and scoring metric choice \citep{liu2024}. A
systematic review of 84 academic and industry papers found that technical performance
was represented in 83\% of evaluated works, while human-centered evaluation appeared
in only 30\%, and both were combined in only 15\% \citep{meimandi2025}. Benchmark
success and deployment value remain systematically disconnected.

Content validity failures are measurable and consequential. A systematic analysis of
194,955 benchmark questions mapped against the EU~AI~Act's taxonomy of model
capabilities found that current benchmarks devote 61.6\% of regulatory-relevant
coverage to hallucination tendency and 31.2\% to performance reliability, while
capabilities central to loss-of-control scenarios (including evading human oversight,
self-replication, and autonomous AI development) receive zero coverage across the
entire public benchmark corpus \citep{prandi2025}.

\subsection{Inter-Rater Reliability: Metrics and Their Conditions of Use}
\label{sec:irr-metrics}

Inter-rater reliability (IRR) quantifies the degree to which raters (human
annotators, LLM judges, or Agent-as-a-Judge systems) agree on the same outcome.
The appropriate IRR metric is determined by the structure of the rating design, not
by convention or familiarity.

\textbf{\ck} \citep{cohen1960} is appropriate when exactly two fixed raters score
every item using the same two-rater configuration across all items. It corrects for
chance agreement using per-rater marginal distributions.

\textbf{\fk} \citep{fleiss1971} generalizes to three or more raters, with the
critical property that rater identity may vary per item provided the number of
ratings per item is constant. It uses the pooled distribution across all raters for
its chance-agreement correction. This is the correct metric for fixed-size judge
panels where individual judge identity may rotate.

\textbf{\ka} \citep{krippendorff2004} is the most general metric: it handles any
number of raters with varying identity, supports missing data natively, and
accommodates nominal, ordinal, interval, and continuous measurement scales through
interchangeable distance functions. For ordinal rubrics, it uses ranked distance
rather than binary mismatch, correctly weighting the severity of disagreement.

\textbf{Intraclass Correlation Coefficient (ICC)} extends IRR concepts to
continuous outcomes. The two-way mixed-effects absolute agreement form,
$\text{ICC}(A,1)$, measures the degree to which scores from one rater (or
measurement occasion) agree in absolute value with scores from another. Unlike
$\kappa$-family metrics, $\text{ICC}(A,1)$ is appropriate when outcomes are
measured on a continuous or interval scale and when absolute agreement (not merely
consistency of ranking) is the validity target. In the context of this paper, it
is used to quantify simulation calibration: the agreement between agent success
rates under simulated versus real user conditions ($V_2$, Section~\ref{sec:layer2}).

Table~\ref{tab:irr-selection} summarizes metric selection by scenario.

\begin{table}[htbp]
\centering
\caption{IRR metric selection by agentic evaluation scenario.}
\label{tab:irr-selection}
\small
\begin{tabular}{@{}p{0.38\linewidth}p{0.25\linewidth}p{0.28\linewidth}@{}}
\toprule
Scenario & Recommended metric & Reason \\
\midrule
3+ LLM judges, binary, complete matrix
  & \fk & Nominal data, complete, multi-rater \\[4pt]
3+ LLM judges, 1--5 rubric, complete matrix
  & \ka{} (ordinal) & Ordinal distance metric needed \\[4pt]
3+ LLM judges, continuous scores, complete matrix
  & \ka{} (interval) & Interval distance required \\[4pt]
2 fixed raters, categorical labels, every item
  & \ck & The two-rater nominal case, and the only one \ck{} fits \\[4pt]
2 fixed raters, ordinal labels, every item
  & weighted \ck{} or \ka{} (ordinal) & Two raters, but an ordinal scale needs a
    distance function; unweighted \ck{} here is the Mode~2 error of
    \S\ref{sec:scan} \\[4pt]
2 fixed raters, continuous scores, every item
  & $\text{ICC}(A,1)$ & Two-rater absolute agreement on a continuous scale \\[4pt]
Rotating judge panel, constant number of judges per item
  & \fk{} (nominal) or \ka{} by scale & Identity may rotate; what matters is that
    the count per item is constant \\[4pt]
Ratings missing, or the number of raters per item itself varies
  & \ka{} (general) & The only design requiring the general form \\
\bottomrule
\end{tabular}
\end{table}

\subsection{Reliability Thresholds and Their Origins}
\label{sec:thresholds}

The Landis--Koch scale \citep{landis1977} is the most frequently cited threshold
system in AI evaluation: $\kappa > 0.21$ as ``fair,'' $\kappa > 0.41$ as
``moderate,'' $\kappa > 0.61$ as ``substantial.'' This scale was developed for
behavioral science contexts with human raters and has no special authority in AI
evaluation. Its widespread use as an acceptability benchmark for LLM judge agreement
is a category error.

For agentic evaluation, domain-calibrated thresholds grounded in construct validity
theory are more appropriate. We propose the following tiered thresholds, derived in
Section~\ref{sec:prescriptions}:

\begin{itemize}
  \item \textbf{Exploratory capability evaluation}: $\alpha \geq 0.67$
    \citep[pp.~241--243]{krippendorff2004}
  \item \textbf{Safety, bias, fairness, and risk scoring}: $\alpha \geq 0.70$
    (\emph{proposed; see Prescription~5})
  \item \textbf{Scoring that gates deployment, compliance, or regulatory reporting}:
    $\alpha \geq 0.80$ (\emph{proposed; see Prescription~5})
\end{itemize}

% -----------------------------------------------------------------------
\section{Related Work}
\label{sec:related}

\subsection{Evaluation Documentation and Knowledge Frameworks}

Structured knowledge frameworks for AI evaluation documentation have emerged as a
response to inconsistent reporting practices. \citet{mitchell2019} introduced Model
Cards, encoding expert knowledge about model behavior, intended use, and limitations
into a standardized artifact that practitioners can apply without deep ML expertise.
\citet{dhar2025} extended this to \textsc{EvalCards}, a structured template for
evaluation-methodology disclosure organized around reproducibility, accessibility,
and accountability. It does not yet require an inter-rater reliability field, which
is precisely the addition Prescription~8 proposes. These frameworks follow the knowledge-based systems
tradition of encoding expert domain knowledge into structured decision procedures that
non-specialists can apply consistently \citep{hayes1983,buchanan1984}: just as early
medical expert systems (MYCIN, INTERNIST) encoded diagnostic knowledge into rule-based
structures, documentation frameworks encode evaluation methodology expertise into
required reporting fields. The compounding validity model ($\vtot \leq V_1 \times V_2
\times V_3$) and the IRR metric selection decision tree proposed in this paper
complement these documentation standards by providing the underlying validity model
that determines \emph{what} to report before documentation is written.

\subsection{Validity Theory Applied to AI Evaluation}

The foundational measurement science concept this paper applies is construct
validity \citep{cronbach1955}: whether a measurement instrument captures the
phenomenon it claims to measure. Despite its centrality, construct validity has
received limited attention in AI evaluation practice \citep{liu2024}.
\citet{jacobs2021} demonstrated that fairness metrics routinely measure constructs
different from their stated targets, a specific instance of the construct validity
failure this paper characterizes at the pipeline level. \citet{meimandi2025} provided
empirical evidence that agentic AI evaluation systematically fails to support the
productivity claims it is used to justify. That result is about which dimensions
are measured rather than about task-distribution fidelity, so it bears on the
content-validity argument of Section~\ref{sec:construct-validity} rather than on
$V_1$ directly. The three-layer compounding model
operationalizes construct validity theory for multi-stage automated pipelines,
providing a tractable computational framework for what has previously been treated
as a qualitative concern.

\subsection{Decision Support for Reliability Metric Selection}

The IRR metric selection decision tree (Figure~\ref{fig:decision-tree}) follows the
expert system tradition of encoding domain expertise into formal decision procedures
that practitioners can apply without deep methodological background
\citep{krippendorff2004}. \citet{james2026} provided a systematic guide to
inter-annotator agreement metric selection in NLP annotation, arguing against
one-size-fits-all use of Cohen's $\kappa$, a position directly aligned with the
prescriptions in Section~\ref{sec:prescriptions}. The decision tree proposed here
extends this guidance to agentic AI evaluation pipelines, encoding the structural
conditions (rater count, identity stability, measurement scale) that determine metric
validity into a formal procedure. This knowledge representation is the framework's
primary deployable artifact: it transforms implicit expert knowledge about IRR metric
selection into an explicit, auditable decision structure.

% -----------------------------------------------------------------------
\section{Layer 1: Task Generation Validity}
\label{sec:layer1}

In traditional static benchmarks, tasks and ground truth are fixed. Reliability can
be validated once against a stable reference. In agentic evaluation, tasks are often
dynamically generated by language models, embedding systematic biases from the
generator into the evaluation distribution before a single agent response is scored.

The construct validity problem at this layer is that dynamically generated tasks may
not sample the intended construct uniformly. An LLM task generator that
overrepresents certain task types, difficulty levels, or surface forms creates a
distribution that raters can agree on perfectly, while measuring something
systematically different from the intended capability.

The empirical evidence is specific. A preprint study (under review at time of
submission) evaluating 10 popular agentic benchmarks found task validity flaws in
7 of them and outcome validity flaws in 7 of them, with all 10 showing benchmark
reporting gaps \citep{zhu2025}. Documented failures include
task validity failures where do-nothing agents passed 38\% of airline booking tasks,
and outcome validity failures where LLM judges made arithmetic errors. Every benchmark
showed at least one validity failure category; none provided sufficient evidence to
support unqualified capability claims.

Task generation validity must be assessed before any IRR analysis is meaningful. A
rating matrix derived from systematically biased tasks cannot be rescued by metric
sophistication at the judgment layer.

% -----------------------------------------------------------------------
\section{Layer 2: Simulation Calibration}
\label{sec:layer2}

Many agentic benchmarks evaluate agents in interaction with simulated users or
simulated world environments rather than real humans and real environments. This
design choice is operationally sensible (real human interaction at evaluation scale
is expensive and slow), but it introduces a second, distinct validity problem:
whether the simulated environment faithfully proxies the real deployment context.

\subsection{Empirical Evidence of Simulation Miscalibration}
\label{sec:sim-evidence}

The most direct empirical evidence comes from \citet{seshadri2026}, which tested
LLM user simulation against real human users on $\tau$-Bench retail tasks across
participants in the United States, India, Kenya, and Nigeria. The findings establish
simulation miscalibration as a measured fact rather than a theoretical concern:

\begin{itemize}
  \item Agent success rates varied up to 9 percentage points across different LLM
    user simulators, demonstrating that simulation results are not robust to
    simulator choice.
  \item Evaluations using simulated users exhibited systematic directional
    miscalibration: underestimating agent performance on challenging tasks while
    overestimating it on moderately difficult ones.
  \item Simulated users introduced conversational artifacts absent from real
    interactions: elevated question-asking, heightened politeness markers, and
    artificial turn structures.
\end{itemize}

\subsection{The Demographic Asymmetry Problem}
\label{sec:demographic}

The calibration failures documented by \citet{seshadri2026} are not uniformly
distributed across user populations. African American Vernacular English (AAVE)
speakers experienced consistently worse success rates and calibration errors than
Standard American English (SAE) speakers, with disparities compounding with age.
Indian English speakers showed similar patterns.

This is simultaneously a measurement validity failure and a fairness failure. An
evaluation pipeline calibrated on SAE simulated users is not a valid measurement of
agentic capability for AAVE-speaking or non-SAE users. As shown in Section~\ref{sec:layer3},
the same demographic asymmetry is compounded at the judgment layer by LLM judges that
exhibit systematic miscalibration across dialect groups independently of simulation
layer failures.

\subsection{Why Agreement on a Distorted Signal Does Not Establish Measurement Validity}
\label{sec:distorted-signal}

The key inferential step at this layer is often missed: high rater agreement on
simulated interactions does not demonstrate evaluation reliability. It demonstrates
that all raters agree on a distorted input.

A scoring panel that perfectly agrees on agent scores produced by an
AAVE-miscalibrated simulator has achieved high IRR in the technical sense. It has
also produced a high-reliability measurement of the wrong thing. Agreement on a
distorted signal is agreement on distortion.

This distinction matters for how $\kappa$ is used as a $V_3$ proxy throughout this
paper (Section~\ref{sec:worked-example}). Where an external human criterion is
available, $\kappa$ between a grader and that criterion measures the grader's
\emph{consistency} with an independently established reference, not correctness in
an absolute sense; this is the standard, imperfect way the field operationalizes
grader validity, and it is what the $V_3$ estimates in Table~\ref{tab:vtotal} are
intended to capture. This is a different measurement target from the raters-agreeing-
on-a-distorted-input problem above, where no external criterion is involved and
agreement is diagnostic of nothing beyond the raters' shared exposure to the same
distortion.

% -----------------------------------------------------------------------
\section{Layer 3: Metric Misspecification}
\label{sec:layer3}

The third layer of validity failure occurs at judgment: even when a human or LLM
judge scores agent outputs, the inter-rater reliability metric used to validate that
scoring is structurally mismatched to the pipeline design in the large majority of
published evaluations.

\subsection{Why \ck{} Is Structurally Wrong for Most Agentic Evaluation Designs}
\label{sec:cohen-kappa}

\ck{} requires exactly two raters scoring every item, with the same two-rater pair
maintained across all items. Almost no agentic evaluation design satisfies this
constraint:

\begin{itemize}
  \item LLM-as-a-Judge pipelines typically use a rotating pool of judge models that
    vary across evaluation runs or item subsets.
  \item Agent-as-a-Judge frameworks \citep{zhuge2025} replace the rater with a
    multi-component agentic pipeline whose judgment is assembled from several
    interacting modules, so ``rater identity'' is not a fixed quantity that a
    two-rater chance correction can condition on.
  \item Human annotation at scale uses crowdsourcing pools where rater identity
    varies per item.
  \item Dynamic benchmark generation produces item batches scored by different
    annotator subsets.
\end{itemize}

Applying \ck{} to any of these designs produces a mathematically invalid result,
because its chance-agreement correction assumes fixed rater identity. Using it
anyway (common in practice) is a silent validity failure that produces an
uncorrectable reliability estimate without warning.

\subsection{The Literature Scan: IRR in Published Agentic Evaluation Papers}
\label{sec:scan}

\textit{Methodology.} We conducted a structured survey of 55 papers published or
posted between 2022 and 2026, selected via a stratified purposive approach across
nine topic categories: major agentic benchmarks; automated grader validity studies;
benchmarks with correct IRR use; LLM-as-a-Judge studies; benchmarks with structural
metric mismatch; long-horizon evaluation; safety and RLHF annotation; recent
2025--2026 evaluation papers; and safety-critical IRR failure cases. Within each
category, papers were selected for citation prominence, venue tier (ACL, EMNLP,
NeurIPS, ICLR), and direct relevance to evaluation methodology; arXiv preprints in
cs.AI and cs.CL supplemented peer-reviewed coverage where thin. This is a purposive
stratified survey, not a systematic review: the sample is designed to cover the main
failure modes rather than to exhaust a keyword-defined corpus. The complete
per-paper coding table is given in \ref{app:literature}
(Table~\ref{tab:scan-full}).

Coding was applied against four pre-specified dimensions: (1)~the IRR metric reported
or absent, (2)~the rater design (number of raters, fixed or rotating identity),
(3)~the measurement scale (binary, nominal, ordinal, or continuous), and (4)~whether
the metric's structural assumptions were satisfied by the pipeline design.
The full coding criteria are documented in \ref{app:literature}.

\textit{Reliability validation.} Coding was performed by the primary author. To
assess coding reliability, all 55 papers in the corpus were coded independently
by three LLM raters from
distinct model families: NVIDIA Nemotron-Ultra-550B, Google Gemma-4-31B, and
Alibaba Qwen3-80B-A3B \citep{openrouter2026}. LLM raters were chosen
deliberately rather than as a substitute for an unavailable human coder:
LLM-as-a-Judge is the prevailing rating pattern in commercial and academic
benchmarks, and it is precisely the practice whose reliability disclosure this
paper argues is inadequate. The check therefore applies the paper's own
prescriptions to that practice: a cross-family panel (Prescription~4), the
structurally correct metric for a four-rater, complete-matrix nominal design
(Fleiss'~$\kappa$, Rule~R3 of Table~\ref{tab:production-rules}), reported
alongside Krippendorff's~$\alpha$ (nominal), and full reporting of the result
including pairwise breakdowns (Prescription~8). We lead with $\alpha$ for
comparability with the thresholds of Section~\ref{sec:thresholds} and report the
R3-prescribed $\kappa$ beside it; on a complete matrix of this shape the two
coincide to within 0.001, which is itself a check that neither was
mis-specified. Each LLM rater received the
coding instrument together with the full extracted text of the paper under
coding, with no access to the primary-rater codes. The stimulus is the entire
document rather than a curated excerpt, so that locating the relevant evidence
is part of the coding task rather than a step already performed for the rater.
Each rater coded all 55 papers in five independent runs, and the author's codes
cover all 55 papers, so the design is a genuinely complete four-rater matrix.

Because each rater was asked to quote the sentence its code rested on, the
returned evidence can itself be checked against the stimulus. A released probe
(\texttt{validate.py}) looks for each quoted run of words in the document the
rater was coding. It locates 89\% of the quotes for the metric-identification
dimension, 85\% for rater design and 88\% for measurement scale; a further
8--10\% cannot be found in the source at all. Those are codes delivered with
evidence that does not exist in the document, and they are reported here rather
than filtered out, because a rater that supplies a confident code attached to an
unlocatable quotation is a sharper instance of this paper's argument than any of
the coefficients below.

Reliability differs sharply by what is being coded. On the
metric-identification dimension (Q1: which coefficient, if any, the authors
computed on their own
annotations), four-way agreement is Krippendorff's $\alpha = 0.834$ (median
across the five runs; range $0.818$--$0.846$), with Fleiss'
$\kappa = 0.818$--$0.846$, at or above the $\alpha \geq 0.80$ level this paper
proposes for deployment-gating use. On the three interpretive dimensions it is
substantially lower: $\alpha = 0.371$ for rater design (Q2), $\alpha = 0.525$ for
measurement scale (Q3), and $\alpha = 0.536$ for the structural-validity
judgment itself (Q4), the dimension that assigns each paper a failure mode.
Collapsing the failure-mode code to the binary distinction that generates this
paper's headline figure, correct use versus misuse, gives $\alpha = 0.549$,
with the four raters placing the misuse rate between 60.7\% and 84.4\% around
the author's 80.0\%. Within-rater stability across the five runs was 100\%
(Gemma), 98\% (Qwen), and 95\% (Nemotron) on the metric-identification
dimension, so the between-rater figures are not an artifact of sampling noise. That
stability is markedly higher than \citet{haldar2025rating} report for LLM judges
scoring open-ended generation, where agreement of a judge with itself across
repeated runs is low enough that they describe the resulting ratings as almost
arbitrary in the worst case. The contrast supports the division this section documents: a judge
asked which coefficient a paper computed is answering a question whose answer is
printed in the paper, whereas a judge scoring generation quality is supplying the
standard as well as applying it. \citet{wei2025systematic} reach the same
separation from a different direction: they model a judge's run-to-run variation
as a flip probability, report it as a self-consistent rate of 0.91--0.98 across
temperature settings and both of their datasets, and then remove it from their
agreement estimate rather than letting it inflate one. The principle that within-rater instability and between-rater disagreement
are different quantities is therefore not particular to this study.

This division is informative rather than incidental. Determinate questions,
whose answer is printed somewhere in the paper, are coded reliably; judgments
requiring a coder to decide whether a rating design satisfies a metric's
structural assumptions are not. The pairwise breakdown shows the same pattern:
on metric identification (Q1) the author agrees with Gemma at $\alpha = 0.924$,
Nemotron at $0.873$, and Qwen at $0.750$, whereas on the structural-validity
judgment (Q4) the corresponding values fall to $0.731$, $0.604$, and $0.412$. We therefore
report the 80\% misuse rate with its reproducibility range attached rather than
as a point estimate, and Section~\ref{sec:internal-threats} discusses what the
residual disagreement reveals. The pairwise matrices for the other two
dimensions are released with the per-dimension reports rather than reproduced
here; on rater design in particular the author and Qwen agree at
$\alpha = 0.029$, which is the weakest cell anywhere in the study.

\textit{Finding.} The most securely established result is the simplest one: 26
of the 55 papers, 47\%, report no chance-corrected inter-rater statistic of any
kind. That is a determination about which coefficient appears in a paper, the
dimension four independent raters recover at $\alpha = 0.834$, and it does not
depend on anyone agreeing about what counts as misuse. Every claim this paper
makes about the judgment layer follows from it alone.

The broader characterization is that 11 papers (20\%) apply a structurally
correct IRR metric with a stated threshold and rationale, and the remaining 44
(80\%) fall into one of four failure modes. These follow the structural-validity
code of the coding instrument, with its \textsc{inc} code separated into
percent-agreement-only and correct-coefficient-without-threshold cases
(\ref{app:literature}, Table~\ref{tab:scan-full}). This 80\% figure carries the
reliability caveat established above, and two further qualifications belong with
it. First, five of the 55 conduct no rating or annotation study of their own, so
there is nothing for which they could have reported reliability: a metric
selection guide, a benchmark checklist, a survey, and two theoretical papers.
Restricting the rate to the 50 papers that do rate something gives 78.0\%
(39 of 50), which is the more conservative reading and the one we would defend.
Second, the categories were designed to span failure modes rather than to sample
the field, which inflates the rate further (Section~\ref{sec:internal-threats}):

\paragraph{Mode 1: No IRR Metric Reported (26 papers, 47\%).}
Within this largest group, two sub-patterns recur. The first is automated grading with no human
reliability measurement: major benchmark papers ($\tau$-bench \citep{yao2024},
\textsc{OSWorld} \citep{xie2024}, \textsc{SWE-bench} \citep{jimenez2024},
\textsc{GAIA} \citep{mialon2024}, and \textsc{WebArena} \citep{zhou2024webarena})
use deterministic automated graders or unvalidated LLM judges with no IRR
measurement against human annotation; $\tau^2$-Bench \citep{barres2025}, the
quality-fix follow-up to $\tau$-bench, takes the same approach, tighter automated
grading rather than human IRR validation. \citet{gurram2026} (preprint, under
review at time of submission) directly measures the gap such graders leave, and
its full set of coefficients is more informative than any one of them:
substring-based automated evaluation reaches $\kappa = 0.049$ against each of
two human annotators, essentially chance level; a three-LLM ensemble reaches
$\kappa = 0.432$; a single GPT-4o-mini judge does better still at
$\kappa = 0.567$; and the two human annotators agree with each other at
$\kappa = 0.835$. Automated grading therefore sits well below the human--human
ceiling even at its best, and the ensemble is not the strongest automated
option, so the gap is not closed simply by adding judges. The Verified re-release of \textsc{WebArena}
(\textsc{WebArena Verified}, \citealt{elhattami2025}: Cohen's $\kappa = 0.83$ on
its task-audit labels; Section~\ref{sec:decision-tree}) shows that rigorous
human annotation produces fundamentally different reliability evidence, even
where the labeled construct is benchmark quality rather than agent outcome. The
second sub-pattern is evaluation that reports a statistic which is not an IRR
metric, or no rater study at all: Pearson correlation for ordinal safety
ratings from 112 demographically diverse annotators \citep{movva2024}; ELO and
Bradley--Terry ratings substituting for inter-annotator reliability in
crowdsourced pairwise preference \citep{chiang2024}; and RLHF annotation
pipelines, red-teaming reports, surveys, and position papers in which the
absence of reliability reporting is treated as unremarkable. In every case the
absence of any reliability report is a complete $V_3$ validity gap.

This sub-pattern is not an artifact of the sampling frame, and it reaches work
written specifically to improve judge reliability.
\citet{wei2025systematic}, which is outside this scan's agentic scope and so is
not one of the 26 papers counted above, sets out to give LLM-as-a-Judge
``more theoretically interpretable evaluation metrics'' and to mitigate judge
internal inconsistency. It reports judge agreement with human evaluators as raw
accuracy, mostly below 0.7, and introduces no chance-corrected coefficient at
any point. A reader cannot tell from an accuracy of 0.68 whether the judge is
tracking the human standard or tracking the marginal distribution of the
labels, which is the distinction a chance correction exists to make. When a
paper devoted to measuring judge reliability still omits that correction, the
omission is better read as a convention of the field than as an oversight by
individual authors.

\paragraph{Mode 2: Structural Metric Mismatch (11 papers, 20\%).}
Papers that do report a chance-corrected metric most often apply one whose
structural assumptions the rating design does not satisfy. Five apply an
unweighted $\kappa$-family statistic to an ordinal or ternary scale without a
distance correction (\citealt{fan2026} AgentProcessBench: unweighted Cohen's
$\kappa = 0.767$ on a ternary $\{-1,0,+1\}$ scale, alongside the commonly-cited
89.1\% raw agreement figure; \citealt{ganguli2022}: Fleiss' $\kappa = 0.32$ on a
five-point Likert scale). Four apply an inter-rater metric (Cohen's $\kappa$,
Fleiss' $\kappa$, Krippendorff's $\alpha$, or ICC) to repeated stochastic runs
of a \emph{single} model, a test--retest self-consistency design in which no
second rater exists and the chance-agreement correction is undefined
(Table~\ref{tab:scan-full}, entries 6, 8, 46, and 47). In \citet{han2025} the
mismatch is more subtle: \ck{} is applied as a series of pairwise comparisons
(each of 54 LLMs vs.\ human ratings), each individually a valid two-rater
design, but the results are then aggregated and interpreted as panel
reliability across 54 configurations with varying rater compositions, a use
case requiring \fk{} or \ka, not repeated two-rater $\kappa$. The same
averaging pattern recurs in a hate-speech reliability study that computes
Cohen's $\kappa$ for all annotator pairs and reports the average across three
annotators drawn from rotating crowdsourced pools
(Table~\ref{tab:scan-full}, entry~18); because rater identity varies across
items, the design calls for \ka{} or \fk{} rather than an averaged two-rater
$\kappa$. That paper explicitly argues that Cohen's $\kappa$ is improper for
subjective labeling tasks and then applies it regardless, which illustrates how
weakly metric choice is currently constrained even by authors alert to the
problem.

What makes both of these a mismatch is that the averaged pairs rate the
\emph{same} items. Averaging pairwise $\kappa$ over a shared item set conflates
variance between rater pairs with agreement about items, which is precisely the
case \fk{} and \ka{} exist to handle. Aggregating per-pair coefficients computed
on \emph{disjoint} item sets is a different operation and is not a mismatch:
\textsc{WebArena Verified} (Section~\ref{sec:decision-tree}) does the latter,
reporting a task-weighted overall $\kappa$ across four site-specific fixed
pairs alongside the per-site values from which it is formed. Each of its
site-level values is in turn a macro-average across annotation categories on
the same items, which is a different averaging axis and not the one at issue
here: the categories are distinct constructs each rated by the same fixed pair,
not competing raters of one construct.

\paragraph{Mode 3: Percent Agreement Only, No Chance Correction (5 papers, 9\%).}
The remaining papers report raw agreement with no correction for chance: 40
contractors ranking model outputs \citep[72--77\%]{ouyang2022}; a 95.9\%
inter-annotator agreement figure for binary web-agent outcome labels
(Table~\ref{tab:scan-full}, entry~10); an 85\% two-reviewer screening
agreement \citep{meimandi2025}; raw LLM--human accuracy for dialectal toxicity
judgments \citep{faisal2024}; and an approximately 85\% agreement rate between
an LLM judge and human labels for web-agent success (entry~39). Raw agreement
is uninterpretable without a chance baseline: for a binary label with skewed
marginals, high percent agreement is compatible with near-zero $\kappa$.

\paragraph{Mode 4: Correct Coefficient, Incomplete Reporting (2 papers, 4\%).}
Two papers select a structurally appropriate coefficient and then report it
without the threshold or rationale that would make the value interpretable. A
large-scale audit of annotation reporting computes a macro-averaged
Krippendorff's $\alpha = 0.585$ for its own annotation study and calls that
``decent'' agreement, a characterization corresponding to no stated convention:
Krippendorff's own guidance treats $\alpha \geq 0.800$ as the threshold for
reliable conclusions and $0.667$ as the floor for tentative ones, and $0.585$
clears neither (Table~\ref{tab:scan-full}, entry~19). A study of length bias in
preference evaluation reports Fleiss' $\kappa = 0.66$--$0.74$ across five raters
per segment, labels that range ``moderate to substantial'' without citing a
banding scheme, and generalizes a reliability claim from an inter-rater analysis
covering only one of its evaluation settings (entry~43). This is the narrowest
mode in the scan and the most direct instance of the paper's central claim: the
coefficient is the correct one and the arithmetic is sound, yet no interpretive
commitment is attached to the number, so the reported value constrains no
conclusion.

Table~\ref{tab:failure-modes} summarizes the distribution.

\begin{table}[htbp]
\centering
\caption{IRR failure modes across 55 individually coded papers, by the
structural-validity code of the coding instrument (\ref{app:literature}). Only
11 (20\%) use structurally correct metrics with explicit rationale.}
\label{tab:failure-modes}
\small
\begin{tabular}{@{}p{0.30\linewidth}rr>{\raggedright\arraybackslash}p{0.42\linewidth}@{}}
\toprule
Failure mode & Papers & \% & Representative example \\
\midrule
No IRR metric reported            & 26 & 47 & $\tau$-bench, \textsc{OSWorld}, \textsc{SWE-bench} (automated grading); Chatbot Arena (ELO); RLHF and red-teaming reports \\
Structural metric mismatch        & 11 & 20 & \ck{} for 54-LLM panel \citep{han2025}; averaged pairwise \ck{} over rotating annotators (entry~18); Fleiss' $\kappa$ on a Likert scale \citep{ganguli2022}; $\alpha$ on repeated runs of one model \\
Percent agreement only            & 5  & 9  & InstructGPT (72--77\%); Emergence WebVoyager (95.9\%) \\
Correct coefficient, incomplete reporting & 2 & 4 & \ka{} $= 0.585$ called ``decent'' with no stated convention (entry~19); \fk{} $= 0.66$--$0.74$ banded without citation (entry~43) \\
\textbf{Correct metric}           & \textbf{11} & \textbf{20} & \textsc{WebArena Verified} ($\kappa = 0.83$, on benchmark-audit labels rather than agent outcomes); MCP-AgentBench (\fk{} = 0.671) \\
\bottomrule
\end{tabular}
\end{table}

\textit{Structural contrast.} The 11 papers using IRR correctly share a
distinguishing behavior: they state the rationale for the metric chosen, and
several report more than one coefficient. Papers coded as mismatched typically
report a single coefficient with no structural justification. The mismatch does
not read as a deliberate simplification; it reflects unawareness of the
structural conditions that govern metric validity.

\subsection{Demographic Asymmetry in LLM Judge Calibration}
\label{sec:judge-demographic}

The metric mismatch problem documented above is compounded by a systematic calibration
asymmetry in LLM judges across linguistic varieties. In a study evaluating LLMs
as toxicity judges across 60 language varieties spanning 10 language clusters,
LLM--human agreement was the weakest dimension of consistency (weaker than
cross-model or cross-dialect consistency), with systematic gaps across dialectal
groups \citep{faisal2024}. Judge miscalibration for non-standard dialect speakers
persists in direct LLM judgment, independently of the simulation layer failures
documented in Section~\ref{sec:demographic}.

The implication for the compounding model is asymmetric $V_3$: benchmarks using LLM
judges without dialect-stratified calibration validation will produce $V_3$ values
that overstate validity for SAE-speaker user populations and understate it for
non-SAE populations. This is an IRR validity failure that structurally correct metric
selection cannot by itself correct, because the miscalibration occurs within the
judge's learned behavior, not in the choice of metric.

\subsection{Ordinal Rubrics and the Kappa Penalty}
\label{sec:kappa-penalty}

Most agentic evaluation rubrics use ordinal scoring (e.g., 1--5 on helpfulness,
safety, coherence). \fk{} treats all disagreements as categorically equal: a rater
scoring 1 when others score 5 is penalized identically to a rater scoring 4 when
others score 5. This is wrong for ordinal data.

The consequence is bidirectional distortion: \ck{} and \fk{} artificially inflate
apparent disagreement when raters are closely aligned on an ordinal scale (one point
apart) and can underestimate disagreement when raters are at opposite ends of the
scale. For safety evaluation, this matters: a safety rubric where judges disagree by
1 point appears as unreliable as one where they disagree by 4 points.

\subsection{Metric Selection Decision Tree}
\label{sec:decision-tree}

Figure~\ref{fig:decision-tree} provides a decision tree for selecting the
structurally correct IRR metric based on rater design, data completeness, and
measurement scale.

\begin{figure}[htbp]
\centering
\begin{minipage}[t]{0.65\linewidth}%
\resizebox{\linewidth}{!}{%
\begin{tikzpicture}[
  baseline=(CK.north),
  dec/.style={diamond, draw, thick, fill=blue!6, text width=2.2cm,
              align=center, inner sep=1pt, aspect=1.35, font=\footnotesize},
  term/.style={rectangle, draw, thick, fill=green!8, rounded corners=3pt,
               text width=2.4cm, align=center, minimum height=0.9cm,
               inner sep=6pt, font=\footnotesize},
  arr/.style={-{Latex[length=3mm]}, thick},
  lbl/.style={font=\scriptsize\itshape, fill=white, inner sep=1.5pt}
]
\node[dec] (Q1) at (0,   0) {2 fixed raters, same pair?};
\node[dec] (Q2) at (4.2, 0) {Missing ratings, or raters per item varies?};
\node[dec] (Q3) at (8.4, 0) {Rubric type?};
\node[term, text width=3.0cm, minimum height=1.7cm] (CK) at (0, 3.4) {%
  \textbf{2-rater, same pair}\\
  {\scriptsize Nominal/binary: Cohen's $\kappa$}\\
  {\scriptsize Ordinal: weighted $\kappa$ / Krippendorff's $\alpha$}\\
  {\scriptsize Interval/continuous: $\text{ICC}(A,1)$}};
\node[term] (KAgen) at (4.2, 2.5) {\textbf{Krippendorff's $\alpha$}\\{\scriptsize general form}};
\node[term] (FK)    at (11.6,  1.4) {\textbf{Fleiss' $\kappa$}\\{\scriptsize nominal, constant count}};
\node[term] (KAord) at (11.6, -1.4) {\textbf{Krippendorff's $\alpha$}\\{\scriptsize ordinal distance}};
\draw[arr] (Q1.east) -- node[lbl, below=2pt]{NO} (Q2.west);
\draw[arr] (Q2.east) -- node[lbl, below=2pt]{NO} (Q3.west);
\draw[arr] (Q1.north) -- node[lbl, right=2pt]{YES} (CK.south);
\draw[arr] (Q2.north) -- node[lbl, right=2pt]{YES} (KAgen.south);
\coordinate (forkX) at (FK.south |- Q3.east);
\draw[thick] (Q3.east) -- (forkX);
\draw[arr]   (forkX)   -- (FK.south);
\draw[arr]   (forkX)   -- (KAord.north);
\node[lbl, above=3pt, align=center] at (FK.north)    {Nominal/binary};
\node[lbl, below=3pt, align=center] at (KAord.south) {Ordinal/interval/continuous};
\end{tikzpicture}%
}%
\end{minipage}%
\hfill%
\begin{minipage}[t]{0.31\linewidth}%
\scriptsize
\textbf{Selection rules:}
\medskip

\noindent\textbf{Cohen's $\kappa$}\\
Exactly 2 fixed raters, same pair, every item, nominal/binary.
\medskip

\noindent\textbf{Weighted $\kappa$ / Krippendorff's $\alpha$}\\
Exactly 2 fixed raters, same pair, every item, ordinal.
\medskip

\noindent\textbf{$\text{ICC}(A,1)$}\\
Exactly 2 fixed raters, same pair, every item, interval/continuous.
\medskip

\noindent\textbf{Fleiss' $\kappa$}\\
$\geq$3 raters (fixed or rotating), constant count per item, nominal/binary.
\medskip

\noindent\textbf{Krippendorff's $\alpha$} (general)\\
Missing ratings, or the number of raters per item itself varies.
\medskip

\noindent\textbf{Krippendorff's $\alpha$} (ordinal/interval)\\
$\geq$3 raters (fixed or rotating), constant count per item, ordinal/interval/continuous.
\medskip

{\tiny\itshape Using \ck{} outside the 2-rater
fixed-pair case produces an invalid reliability
estimate without warning.}
\end{minipage}
\caption{IRR metric selection. Left: decision tree based on rater design, matrix
completeness, and rubric scale. Right: selection rules summary. A rotating panel
with a constant number of raters per item (e.g.\ a fixed-size rotating judge pool)
follows the same path as a fixed panel of the same size (Fleiss' $\kappa$ or
Krippendorff's $\alpha$, \S\ref{sec:irr-metrics}); Krippendorff's $\alpha$ (general)
is reserved for designs where that count is itself irregular or ratings are
missing. Using \ck{} outside the two-rater fixed-pair case is a silent validity
failure.}
\label{fig:decision-tree}
\end{figure}

Figure~\ref{fig:decision-tree} is a knowledge-based decision support system in the
expert systems tradition~\citep{hayes1983,buchanan1984,james2026}: it encodes domain
expert knowledge about IRR metric structural assumptions into a formal decision
procedure that practitioners can apply without background in psychometrics. The
terminal nodes are executable recommendations corresponding to the six production
rules given as Table~\ref{tab:production-rules} in \ref{app:literature},
\S{}A.5; the two-rater terminal resolves into three rules by measurement scale
(R1, R1$^\dagger$, R1$^\ddagger$). A practitioner who provides the rater
count, identity stability, and scale type obtains a structurally correct metric
recommendation without needing to understand the mathematical foundations of
chance-agreement correction. This procedural knowledge representation is the
framework's primary deployable artifact, complementing the governance prescriptions
in Section~\ref{sec:prescriptions}.

\textbf{Worked example 1 (LLM-as-judge, ordinal rubric).}
An evaluation uses three LLM judges (GPT, Claude, Gemini) scoring agent responses on
a 1--5 helpfulness rubric, with all judges scoring every item. Applying the rules:
rater\_count = 3 (not R1); complete matrix, no missing ratings (not R2); scale is
ordinal (not R3). Rule R4 applies: Krippendorff's~$\alpha$ with ordinal distance
function. Reporting \ck{} for this design (a common practice) violates R1's
two-rater constraint and produces an invalid estimate.

\textbf{Worked example 2 (human annotation, binary labels).}
\textsc{WebArena Verified} \citep{elhattami2025} assigns one primary annotator
per site across the four self-hosted sites (Map is excluded from their
reliability analysis because it is not self-hosted) and re-labels 100\% of tasks
with a paired blind verifier (site pairs A$\leftrightarrow$B,
C$\leftrightarrow$D), so each task is scored by a fixed pair on a binary
label. Applying the rules:
rater\_count = 2 (per pair), same pair for every item in that pair's sites,
complete matrix (R1). Cohen's $\kappa$ is the correct metric, and the reported
site-level values (0.81--0.85; task-weighted overall $\kappa = 0.83$, 95\% CI
[0.81, 0.85]) meet the
deployment-gating threshold ($\alpha \geq 0.80$, Prescription~5) for the
construct actually being rated. The thresholds of
Section~\ref{sec:thresholds} are stated for $\alpha$ and read here against a
$\kappa$; for a binary label on a complete two-rater matrix the two
coefficients are close enough for that comparison to be informative, but it is
an approximation and not an identity, and we flag it rather than let the
substitution pass silently. That construct is task-specification ambiguity
and category-validity labeling, not agent-outcome success or failure; this
example illustrates a structurally correct application of Cohen's $\kappa$
(Prescription~2), but the resulting $\kappa$ cannot be substituted for $V_3$ in
Table~\ref{tab:vtotal}, since $V_3$ requires reliability on agent-outcome
judgment specifically (Section~\ref{sec:worked-example}).

% -----------------------------------------------------------------------
\section{The Compounding Argument}
\label{sec:compounding}

The full mathematical treatment, including the same-provider estimator-bias
discussion and a practical estimation protocol, is in Section~\ref{sec:applying}
and \ref{app:compounding}.

\subsection{Heuristic Validity Bound}
\label{sec:formal}

Let $\cone \in [0,1]$ denote the task generation validity: the degree to which the
generated task distribution matches the ideal construct distribution, defined as one
minus the total variation distance between them. Because the ideal construct
distribution $D_{\mathcal{C}^*}$ is a theoretical object and not directly observable,
$\cone$ cannot be computed from first principles; in practice it is estimated by
comparing generated tasks against a curated expert reference set
(\ref{app:compounding}, \S{}B.6). Let $\ctwo \in [0,1]$ denote the
simulation calibration validity, defined as the $\icc$ between simulated and real
agent outcomes on a matched task set, estimated separately per user population group.
Let $\cthree \in [0,1]$ denote the judgment validity: the appropriate IRR metric
value (\ka{} or \fk, per the pipeline's measurement scale) for the rating design.
As an agreement coefficient, $\cthree$ measures the grader's consistency with an
independently established reference rather than correctness in an absolute sense
(Section~\ref{sec:distorted-signal}); it is a proxy for judgment validity, not a
direct measurement of it.

The overall construct validity of the evaluation pipeline is bounded above by:
\begin{equation}
  \vtot \leq \cone \times \ctwo \times \cthree
  \label{eq:bound}
\end{equation}

The bound holds under the assumption of independence between layer failures. When
the same provider family operates at all three layers, shared systematic biases
not captured in the cross-pipeline estimates of $V_1$, $V_2$, and $V_3$ cause
those estimates to overstate true validity for that specific configuration. The
independence bound is therefore optimistic for same-provider pipelines; the
practical $\vtot$ for such configurations may fall below it
(\ref{app:compounding}, \S{}B.4).

Figure~\ref{fig:pipeline} illustrates the three-layer pipeline with its validity
estimates and the compounding formula.

\begin{figure}[t]
\centering
\resizebox{\textwidth}{!}{%
\begin{tikzpicture}[
  layer/.style={rectangle, draw, thick, fill=gray!6, inner sep=4pt, font=\footnotesize},
  score/.style={rectangle, draw, thick, fill=orange!12, text width=2.0cm,
                align=center, minimum height=4.782cm, rounded corners=5pt,
                inner sep=6pt, font=\footnotesize},
  arr/.style={-{Latex[length=3mm, width=2mm]}, thick},
  lbl/.style={font=\scriptsize\itshape, fill=white, inner sep=2pt}
]
\node[layer] (G) {%
  \parbox[t][4.5cm][s]{3.6cm}{%
    \textbf{Task Generator}\\[2pt]%
    \rule{3.6cm}{0.5pt}%
    \vfill%
    \textit{Failure:} construct drift%
    \vfill%
    \rule{3.6cm}{0.3pt}\\[2pt]%
    {\scriptsize task/outcome validity flaws in 7 of 10; all 10 show reporting gaps \citep{zhu2025}}%
    \vfill%
    \rule{3.6cm}{0.3pt}\\[2pt]%
    $V_1 = 0.62$--$0.97$ (Table~\ref{tab:vtotal})%
  }%
};
\node[layer, right=1.3cm of G] (S) {%
  \parbox[t][4.5cm][s]{3.6cm}{%
    \textbf{World/User Simulator}\\[2pt]%
    \rule{3.6cm}{0.5pt}%
    \vfill%
    \textit{Failure:} demographic miscalibration%
    \vfill%
    \rule{3.6cm}{0.3pt}\\[2pt]%
    {\scriptsize 9pp variance across simulators; AAVE underestimation \citep{seshadri2026}}%
    \vfill%
    \rule{3.6cm}{0.3pt}\\[2pt]%
    $V_2 = 0.80$--$1.00$ (Table~\ref{tab:vtotal})%
  }%
};
\node[layer, right=1.3cm of S] (J) {%
  \parbox[t][4.5cm][s]{3.6cm}{%
    \textbf{LLM Judge Panel}\\[2pt]%
    \rule{3.6cm}{0.5pt}%
    \vfill%
    \textit{Failure:} metric mismatch%
    \vfill%
    \rule{3.6cm}{0.3pt}\\[2pt]%
    {\scriptsize 80\% of papers in this purposive sample use invalid or absent IRR (\S\ref{sec:scan})}%
    \vfill%
    \rule{3.6cm}{0.3pt}\\[2pt]%
    $V_3$: undefined/unpublished (Table~\ref{tab:vtotal})%
  }%
};
\node[score, right=1.3cm of J] (SC) {%
  \textbf{Score}\\[4pt]%
  $\vtot \leq V_1 V_2 V_3$\\[4pt]%
  not\\computable\\[2pt]%
  {\scriptsize see Table~\ref{tab:vtotal}}%
};
\draw[arr] (G.east) -- (S.west);
\draw[arr] (S.east) -- (J.west);
\draw[arr] (J.east) -- (SC.west);
\node[lbl, above=4pt] at ($(G.north east)!0.5!(S.north west)$)  {\itshape tasks};
\node[lbl, above=4pt] at ($(S.north east)!0.5!(J.north west)$)  {\itshape outcomes};
\node[lbl, above=4pt] at ($(J.north east)!0.5!(SC.north west)$) {\itshape ratings};
\end{tikzpicture}%
}%
\caption{Three-layer agentic evaluation pipeline. Each layer introduces a
distinct failure; failures compound multiplicatively under the independence
bound ($\vtot \leq V_1 \times V_2 \times V_3$). $V_1$ and $V_2$ are measured for
the benchmarks in Table~\ref{tab:vtotal}; $V_3$ is not, because three use a
deterministic single-rule grader for which reliability is not a well-defined
question, and the fourth has no published grader-vs-human measurement for its
partial LLM-judgment component, so $\vtot$ cannot currently be computed for any
of them. The bound itself is additionally optimistic for same-provider
pipelines where shared systematic biases cause the external estimates to
overstate true validity (\ref{app:compounding}, \S{}B.4).}
\label{fig:pipeline}
\end{figure}

\subsection{Applying the Model to Published Benchmarks}
\label{sec:worked-example}

Table~\ref{tab:vtotal} applies the compounding model to five widely-used agentic
benchmark configurations using values drawn directly from the cited literature.

\textbf{$V_1$ sources}: these are derived from the per-benchmark failure rates
reported by \citet{zhu2025}, using the rule $V_1 = 1 - r$, where $r$ is the
reported rate and the midpoint is taken where a range is given. Three of the
four are scoped more narrowly than the benchmark name suggests, and we state
the scope rather than let the benchmark label carry it: do-nothing agents passed
38\% of $\tau$-bench's airline subset ($V_1 = 0.62$); broken evaluation
selectors caused 28\% performance underestimation in OSWorld's Chrome task
section ($V_1 = 0.72$); augmented test cases changed 41\% of the SWE-bench
Verified leaderboard, against 24.4\% on Lite, and we take the Verified figure
($V_1 = 0.59$).

One of the four needs a stronger caveat. \citet{zhu2025} attribute WebArena's
1.4--5.2\% misestimation to permissive string matching and to an
LLM-as-a-Judge used without validating its accuracy, which under this paper's
own three-layer division is a judgment-layer defect rather than a
task-generation one. We retain $V_1 = 0.97$ for comparability with the other
rows, but the figure should be read as an upper bound on WebArena's
task-generation validity that happens to be constrained by a Layer-3
measurement, not as a measurement of Layer~1; and it sits in the same row whose
$V_3$ we report as unpublished. More generally, \citet{zhu2025} classify the
OSWorld and SWE-bench findings above as outcome-validity rather than
task-validity issues, so only the $\tau$-bench row is sourced from a finding
their taxonomy and ours assign to the same layer. The compounding model's
assumption that the three layers are independently estimable is therefore better
supported for $\tau$-bench than for the other three, and that is a limitation of
the estimates in Table~\ref{tab:vtotal} rather than of the published
measurements they draw on.

\textbf{$V_2$ sources}: OSWorld, SWE-bench, and WebArena use real computer or web
environments without LLM user simulators; $V_2 = 1.00$ for these.
For $\tau$-bench, \citet{seshadri2026} measured Expected Calibration Error
($\mathrm{ECE}_{\mathrm{Human-LLM}}$) between simulated and real user outcomes.
We operationalize $V_2 \approx 1 - \mathrm{ECE}/100$: 11.7\% for Standard American
English speakers ($V_2 = 0.88$) and 20.3\% for AAVE speakers ($V_2 = 0.80$).

\textbf{$V_3$ sources}: $V_3$, as defined in this paper (Section~\ref{sec:formal}),
is an inter-rater reliability metric ($\kappa$ or $\alpha$), and inter-rater
reliability presupposes at least two raters capable of disagreeing. $\tau$-bench,
OSWorld, and SWE-bench each score agent outputs with a single deterministic rule
(database-state comparison, a custom execution script, and unit-test execution
respectively): there is no second rater to disagree with. For these three
benchmarks, $V_3$-as-reliability is not merely unmeasured; it is not a
well-defined quantity under this model. A deterministic rule agrees with itself
on every application, so it is trivially and perfectly self-consistent regardless
of whether the rule correctly operationalizes the construct it is meant to
check. WebArena (original) is mixed: its \texttt{exact\_match} and
\texttt{must\_include} functions, which score the majority of its 812 tasks,
share this same deterministic, no-second-rater property. Its \texttt{fuzzy\_match}
function, used for a subset of information-seeking tasks, is a genuine
single-LLM judgment step, for which reliability is at least conceptually the
right question, though no WebArena-specific grader-vs-human measurement has been
published. \citet{gurram2026}'s $\kappa = 0.049$ substring-matching estimate,
measured on a different benchmark with a different grader implementation, would
have to be borrowed by analogy rather than applied as a direct measurement, and
even that analogy covers only the minority \texttt{fuzzy\_match} portion of
WebArena's tasks. We therefore do not compute $V_3$ or $\vtot$ for any benchmark
in Table~\ref{tab:vtotal}.

This gap is not merely a reporting omission. Deterministic, single-rule graders
can be perfectly reliable while being invalid, and \texttt{must\_include}-style
substring checks illustrate why: a check for the substring ``Paris'' against the
question ``What is the capital of France?'' scores the response ``Paris is not
the capital; Madrid is'' as correct, every time, with zero disagreement possible,
because the rule only tests for lexical presence rather than semantic
correctness. No inter-rater reliability metric would ever surface this failure,
because there is no second rater whose disagreement could flag it; the rule is
reliable and wrong simultaneously. This is the deterministic-grader analogue of
the argument in Section~\ref{sec:distorted-signal}: agreement (or, here, trivial
self-agreement) on a distorted signal is not evidence of validity.

\begin{table}[htbp]
\centering
\caption{$V_1$ and $V_2$ estimates for five published agentic benchmark
configurations, using values derived from the cited literature. $V_3$ and
$\vtot$ are not reported: three benchmarks use a deterministic single-rule
grader for which reliability is not a defined quantity (no second rater),
and the fourth (WebArena original) has no published grader-vs-human
measurement for its partial LLM-judgment component. $V_2^{\dagger}$: real
environment, no LLM user simulator.}
\label{tab:vtotal}
\small
\begin{tabular}{@{}lccc@{}}
\toprule
Benchmark & $V_1$ & $V_2$ & $V_3$ \\
\midrule
$\tau$-bench retail (SAE users) & 0.62 & 0.88 & not applicable (deterministic grader) \\
$\tau$-bench retail (AAVE users) & 0.62 & 0.80 & not applicable (deterministic grader) \\
OSWorld & 0.72 & 1.00$^{\dagger}$ & not applicable (deterministic grader) \\
SWE-bench & 0.59 & 1.00$^{\dagger}$ & not applicable (deterministic grader) \\
WebArena (original) & 0.97 & 1.00$^{\dagger}$ & not published (partial LLM grader) \\
\bottomrule
\end{tabular}
\end{table}

\smallskip
\noindent\textbf{Caveats on derived estimates.} The $V_1$ and $V_2$ values above
are derived estimates, not direct measurements. $V_1$ is estimated from
published benchmark failure rates \citep{zhu2025} rather than direct total
variation distance measurement, because the ideal construct distribution
$D_{\mathcal{C}^*}$ is a theoretical object not directly observable. $V_2$ is
operationalized via Expected Calibration Error rather than $\mathrm{ICC}(A,1)$
(specifically, $V_2 \approx 1 - \mathrm{ECE}_{\mathrm{Human\text{-}LLM}}/100$
from \citealt{seshadri2026}); this approximation holds only when ECE measures
uniform miscalibration, which is not guaranteed. \ref{app:compounding}
(\S{}B.5) works a fully hypothetical numerical example of the compounding
formula for illustration; it is not tied to any specific published benchmark.

Two findings emerge. First, $V_1$ and $V_2$ are independently measurable and
independently show real degradation: task validity flaws in the majority of
surveyed benchmarks (Section~\ref{sec:layer1}) and up to 9 percentage points of
simulator miscalibration with a demographic gradient (Section~\ref{sec:layer2}).
These layers are not the missing piece; they are measured, published, and
ignored by the benchmarks that report scores derived from them.

Second, and more consequentially: of the five benchmark configurations surveyed
here, none has a construct-appropriate, benchmark-specific measurement of
judgment-layer validity. Three have no defined $V_3$ at all under any
IRR-based model, because their graders are deterministic single rules with no
second rater. The fourth has a published grader-vs-human estimate available
only by analogy from an unrelated benchmark, and only for a minority of its
tasks. This is a stronger and more specific claim than ``$V_3$ is usually
low'': for the large majority of widely-used agentic benchmarks, the question
the compounding model poses at the judgment layer is not merely unanswered, it
is not currently answerable with a reliability metric at all, and a different
validity question (does the deterministic rule correctly operationalize the
construct?) applies instead, one this model does not yet formalize.

\subsection{Implications for Published Benchmark Claims}
\label{sec:implications}

When a paper reports that ``Agent X achieves 82\% on [benchmark],'' the epistemic
weight of that claim is conditioned on the validity of the three pipeline layers.
Table~\ref{tab:vtotal} shows that for five widely-used agentic benchmark
configurations, the judgment layer's validity has never been assessed in a form
this model can quantify: three use a deterministic grader for which reliability
is not a defined question, and the fourth's partial LLM-judgment component has
no published grader-vs-human measurement. An 82\% score from any of these
benchmarks does not tell us that the agent achieves 82\% of the intended
capability, because the conversion from raw score to capability estimate
requires an assessed $V_3$ that does not currently exist for any of them.

This does not mean the score is uninformative, nor does it license a bound such
as $\vtot \leq 0.05$: the deterministic benchmarks' $V_3$ is not a well-defined
quantity to substitute into the compounding formula at all, rather than a
known-low one, and asserting a numeric bound here would repeat the same error
this paper identifies elsewhere. The correct response to an unassessed
judgment layer is to say so, not to assume validity by default and not to
manufacture a number where none currently exists. A benchmark whose scores
carry an assessed judgment-layer validity would require, for benchmarks like
$\tau$-bench, OSWorld, and SWE-bench, the construct-validity question this
model does not yet formalize (does the deterministic rule correctly
operationalize the target construct?), and, for WebArena, a grader-vs-human
measurement specific to its own \texttt{fuzzy\_match} implementation rather
than one borrowed from an unrelated benchmark.

% -----------------------------------------------------------------------
\section{Prescriptions for Valid Agentic Evaluation}
\label{sec:prescriptions}

Items supported by existing literature are stated as requirements; items that extend
beyond current literature are marked \emph{we propose}.

\begin{enumerate}

\item \textbf{Validate simulation calibration before trusting evaluation results.}
Use a held-out set of real human interactions to measure the $\icc$ between simulated
and real agent outcomes. If calibration is not validated, the evaluation signal is of
unknown reliability regardless of how precisely the judging rubric is specified
\citep{seshadri2026}. Report the calibration ICC alongside any benchmark result that
uses user simulation. \emph{We propose $\icc \geq 0.70$ as a minimum threshold for
simulation calibration. The value is a round-number convention set just above
\citet{krippendorff2004}'s general content-analysis minimum of 0.67, not a
quantity derived from the simulator literature; what the evidence supports is
the direction, that LLM simulators exhibit systematic behavioral artifacts
absent from trained-human rater contexts \citep{seshadri2026} and so should not
be held to a lower bar than human coders, rather than the specific 3pp
increment.}

\item \textbf{Select the IRR metric based on pipeline structure, not convention.}
Apply \ck{} only when exactly two fixed raters score every item. Apply \fk{} when a
fixed-size panel of three or more raters scores nominal or categorical outcomes on a
complete matrix. Apply \ka{} when the rubric is ordinal or continuous, when missing
ratings are possible, or when rater panel membership varies. Using \ck{} outside its
structural assumptions is a silent validity failure (Section~\ref{sec:layer3}).

\item \textbf{Report both \fk{} and \ka{} for any rubric using ordinal or interval
scales.} When the two metrics diverge, the divergence is itself diagnostic:
$\kappa < \alpha$ suggests raters are frequently one scale point apart rather than at
opposite ends of the range, a much healthier disagreement pattern than the $\kappa$
score alone would imply.

\item \textbf{Use cross-family judge ensembles.} Self-preference and family-level
bias in LLM judges \citep{spiliopoulou2025,wataoka2024} are large enough to distort
comparative agent rankings. Evaluation pipelines should use judges from at least two
distinct provider families and report inter-judge agreement across families.
Single-family judge panels produce spuriously inflated within-family agreement.

\item \textbf{Apply domain-appropriate reliability thresholds.}
For exploratory capability evaluation, $\alpha \geq 0.67$ is the minimum consistent
with \citet[pp.~241--243]{krippendorff2004}'s own recommendation. For safety, bias,
fairness, and risk scoring, \emph{we propose} $\alpha \geq 0.70$. For scoring that
gates model deployment, compliance certification, or regulatory reporting, \emph{we
propose} $\alpha \geq 0.80$. When these thresholds cannot be met, as \citet{jafari2026} (preprint, under review
at time of submission) demonstrate for LLM mental health response evaluation
(three certified psychiatrists produced $\text{ICC} = 0.087$--$0.295$ and
$\alpha = -0.203$), the correct response
is not to lower the threshold but to recognize that the construct is insufficiently
specified for reliable measurement.

\item \textbf{Validate dynamically generated task distributions against curated
reference sets.} When tasks are generated by an LLM, validate that the generated
distribution samples the intended construct by testing a sample of generated tasks
against domain-expert judgment: present tasks to experts who rate each on whether it
faithfully instantiates the target construct, and compute $V_1$ as the proportion
rated as construct-valid \citep{zhu2025}. Where a curated expert reference set
exists, total variation distance between the generated and reference distributions
provides a more formal measure; in practice, expert-rated proportions are the
tractable operationalization (Section~\ref{sec:applying}, Step~1).

\item \textbf{Stratify evaluation by demographic and linguistic group.} Simulation
calibration failures are not uniformly distributed across user populations
\citep{seshadri2026}, and LLM judge calibration failures show the same pattern across
dialect groups \citep{faisal2024}. AAVE speakers, Indian English speakers, and other
non-SAE populations must be included in calibration validation before benchmark
results are considered representative.

\item \textbf{Report IRR as a required field in evaluation documentation.}
Standardized reporting frameworks such as \textsc{EvalCards} \citep{dhar2025} provide
the infrastructure to enforce consistent disclosure across papers and model releases.
Including IRR metric, threshold, and metric-selection rationale as required fields
in evaluation reporting cards would institutionalize the prescriptions above at the
submission level.

\end{enumerate}

% -----------------------------------------------------------------------
\section{Applying the Framework: A Pipeline Assessment Procedure}
\label{sec:applying}

The following procedure operationalizes the compounding validity model as a
knowledge-based pipeline assessment tool. Pipeline operators can apply it at
evaluation design time to estimate $\vtot$ and identify which layer requires the most
attention before results are reported.

\begin{enumerate}

\item \textbf{Estimate $V_1$ (task generation validity).} Generate a sample of 100
  tasks from the pipeline's task generator $G$. Present them to domain experts who
  score each task on whether it faithfully samples the intended construct. Compute
  $V_1$ as the proportion rated as construct-valid. If a curated reference set is
  available, measure total variation distance between generated and reference
  distributions instead.

\item \textbf{Estimate $V_2$ (simulation calibration).} Run a matched subset of
  tasks (minimum 50) with both the LLM simulator $S$ and real human users. Compute
  $\mathrm{ICC}(A,1)$ between simulated and real agent success rates, separately for
  at least two user population groups (e.g., SAE and one non-SAE group). Use the
  population-weighted average as $V_2$.

\item \textbf{Estimate $V_3$ (judgment validity).} Apply the IRR metric selection
  procedure (Figure~\ref{fig:decision-tree}) to identify the structurally correct
  metric for the pipeline's rater design. Compute that metric on a sample of
  ratings against an independent human reference. This value is $V_3$. If the
  judgment layer is a single deterministic rule (database-state comparison,
  unit-test execution, substring matching), $V_3$ as an IRR quantity is
  undefined (Section~\ref{sec:worked-example}); a grader-vs-human validation
  study of the kind reported by \citet{gurram2026} must be conducted for that
  grader before Step~4 can be completed.

\item \textbf{Compute the bound.} $\vtot \leq V_1 \times V_2 \times V_3$. Report
  this bound alongside any benchmark result derived from the pipeline.

\end{enumerate}

\noindent\textbf{Interpretation guideline.} If $\vtot \leq 0.50$, benchmark scores
should not be used for consequential deployment decisions without independent
validation. If $\vtot \leq 0.30$, the pipeline should be redesigned before any
deployment claim is made. The formal definitions and the same-provider
estimator-bias discussion are in \ref{app:compounding}.

% -----------------------------------------------------------------------
\section{Threats to Validity}
\label{sec:threats}

We assess threats to the validity of this study following the construct, internal,
external, and conclusion validity framework standard in empirical research methodology
\citep{wohlin2012,jedlitschka2008}.

\subsection{Construct Validity}
\label{sec:construct-threats}

\textit{Operationalization of $V_1$, $V_2$, $V_3$.} The approximations underlying
each $V_i$ estimate and their implications for Table~\ref{tab:vtotal} are stated
in the ``Caveats on derived estimates'' note directly beneath that table
(Section~\ref{sec:worked-example}). In summary: $V_1$ is estimated from published
benchmark failure rates rather than direct total variation distance; $V_2$ uses
Expected Calibration Error as a proxy for $\mathrm{ICC}(A,1)$, an approximation
valid only under uniform miscalibration; and $V_3$ is not reported for any benchmark in that
table, because it is not a well-defined quantity for a deterministic
single-rule grader and has not been published for the one partial LLM grader
(Section~\ref{sec:worked-example}). Any $\vtot$ computed from inputs of this
kind should be read as a derived bound under these stated assumptions.

\textit{IRR coding taxonomy.} The four-dimension coding scheme
(Section~\ref{sec:scan}) may not capture all relevant failure modes. In particular,
papers that apply the structurally correct metric but with an incorrect threshold, or
that report IRR on a subset of items not representative of the full evaluation, fall
outside the coding scheme and may be classified as ``correct'' when they should not
be. \ref{app:literature} provides the per-paper coding table
(Table~\ref{tab:scan-full}) to support independent assessment.

\textit{Compounding model.}
The multiplicative validity bound (Equation~\ref{eq:bound}) is presented as a
conceptual model, explicitly not a proved theorem (\ref{app:compounding},
\S{}B.3). The motivating argument assumes layer independence and treats each $V_i$
as a normalized signal-to-noise ratio; both assumptions are approximations. The bound
should be read as a formalization of the compounding intuition, not a mathematical
guarantee.

\subsection{Internal Validity}
\label{sec:internal-threats}

\textit{Coding reliability.} The literature scan (Section~\ref{sec:scan}) was coded
by the primary author. To address the potential self-referential tension (a paper about
IRR using single-rater coding), we conducted a four-rater reliability study over
the whole corpus. Three LLMs from distinct model families (NVIDIA
Nemotron-Ultra-550B, Google Gemma-4-31B, Alibaba Qwen3-80B-A3B;
\citealt{openrouter2026}) independently applied the coding instrument to the
complete extracted text of each of the 55 papers, five times each, with no
access to the primary-rater codes. Agreement is high on the
metric-identification dimension (Q1; Krippendorff's $\alpha = 0.834$; Fleiss'
$\kappa = 0.818$--$0.846$) and materially lower on the structural-validity
judgment (Q4) that produces this paper's headline figure ($\alpha = 0.536$, or $0.549$ for the
binary correct-versus-misuse collapse), which falls below the paper's own
exploratory threshold ($\alpha \geq 0.67$). We therefore treat the 80\% IRR
misuse rate as a structured expert estimate carrying a measured reproducibility
range of 60.7\%--84.4\%, rather than as an IRR-validated point estimate. What the
study validates is narrower than the headline claim: independent raters reliably
recover \emph{which} coefficient a paper computed, and agree much less on
\emph{whether} that choice constitutes misuse. This is why
Section~\ref{sec:scan} leads with the count of papers reporting no
chance-corrected coefficient at all, 26 of 55: that quantity rests on the
dimension the study does validate, and the argument of this paper does not need
the contested one.

The same study supplies a second caution about its own raters. On the released
grounding probe, 8--10\% of the quotations the raters offered as the evidence for
a code cannot be located anywhere in the document they were coding
(Section~\ref{sec:scan}). An unlocatable quotation does not by itself make the
code wrong, and the codes were not filtered on it, but it means some fraction of
the agreement measured above was reached without demonstrable grounding in the
stimulus. That is a limitation of this validation exercise and, at the same
time, the most direct evidence in the paper for the practice it criticizes.

The disagreement is not evenly distributed across raters. On the failure-mode
judgment, Gemma agrees with the author at pairwise $\alpha = 0.731$ and Nemotron
at $0.604$, while Qwen3-80B-A3B agrees with the author at only $0.412$ and with
Nemotron at $0.311$. Qwen also departed most often from the instrument's own
stated rule: on 28\% of its no-coefficient codings (36 of 127, affecting 10 of
the 26 papers it coded that way), it failed to record the corresponding
absent-metric failure mode, whereas the other two raters never broke that rule. Because a rater's rule-following rate is
observable only when the instrument states a rule and the codes are reported in
full, this is itself an argument for Prescription~8. The divergence concentrated
on papers with mixed or borderline evidence,
for instance papers reporting a chance-corrected metric alongside a raw
percentage figure, or applying a plausible-sounding metric to a design it does
not quite fit, more than on papers reporting no metric at all. On several of
these borderline papers, Qwen coded the design as structurally correct where the
other three raters coded it as mismatched or incomplete. This is, at a meta
level, an instance of the failure mode this paper documents: a single rater's
calibration, left unreported or averaged away, would silently have shifted the
apparent reliability of the underlying finding. Had we reported only Qwen's
codes, or only a majority vote without the pairwise breakdown, this divergence
would be invisible. We treat its visibility as a direct illustration of
Prescription~8 (IRR as a required reporting field) rather than as noise to be
discarded.

A secondary tension is worth naming explicitly: this paper argues that LLMs exhibit
systematic miscalibration as evaluative raters (Sections~\ref{sec:demographic}
and~\ref{sec:layer3}), yet uses LLM raters as a validation mechanism here. These are
not equivalent roles, and the result above sharpens rather than resolves the
tension: one of the three LLM raters exhibited exactly the kind of miscalibration
the paper warns about elsewhere, detectable here only because reliability was
measured and reported in full. These results do not claim that LLMs are valid
substitutes for domain-expert human raters. The high metric-identification
figure shows only that a determinate question with a printed answer can be coded
reproducibly; the much lower structural-validity figure shows that the interpretive
judgment this paper actually rests on cannot, and that LLM raters diverge from a
human-coded baseline in ways visible only when IRR is computed and disclosed.
That asymmetry is the paper's central argument applied reflexively to its own
methodology. The
pre-specified coding criteria, category summary, and per-paper coding table in
\ref{app:literature} document the coding scheme; \S{}A.4 lists notable
individual cases referenced in the main text.

\textit{Purposive sampling and confirmation bias.} The 55-paper sample was selected
purposively to cover nine topic categories, not to exhaust a keyword-defined corpus.
The 80\% misuse rate is a finding within this purposive sample and should not be
read as a field-wide prevalence estimate. A random sample from a keyword-defined
corpus could produce a materially different rate: papers demonstrating IRR failures
may be more visible and more frequently cited than papers with correct but
underreported IRR practices, which would be underrepresented in a citation-weighted
purposive selection. Additionally, the category design (including categories where
correct IRR use was by design absent, such as Category A) inflates the apparent
misuse rate relative to what a random corpus draw would show. A second source of
inflation is the denominator: five of the 55 papers run no rating or annotation
study of their own and so had nothing to report reliability for, and excluding
them gives 78.0\% (39 of 50), the figure we regard as the more defensible of the
two (Section~\ref{sec:scan}). The structured
category definitions and transparent category membership (\ref{app:literature})
allow readers to assess whether the sampling frame is representative of their own
research context, but do not transform a purposive finding into a prevalence claim.

\subsection{External Validity}
\label{sec:external-threats}

\textit{Sample scope.} The 55 papers were drawn primarily from English-language
venues (ACL, EMNLP, NeurIPS, ICLR) and arXiv preprints in cs.AI and cs.CL. Papers
from other geographic regions, language communities, or non-English venues may show
different IRR reporting patterns. The temporal scope (2022--2026) means that earlier
work and work after the scan cutoff is not captured; IRR practices are evolving and
the field may improve.

\textit{Domain of the compounding model.} The three-layer model was constructed for
agentic evaluation pipelines involving task generation, user simulation, and LLM
judgment. Static benchmarks with fixed tasks, no simulation layer, and human-only
raters present a different threat model where only Layer~3 (IRR metric selection)
directly applies. The prescriptions in Section~\ref{sec:prescriptions} are intended
to generalize across evaluation pipeline designs, but their applicability to static
benchmarks should be assessed separately.

\subsection{Conclusion Validity}
\label{sec:conclusion-threats}

\textit{Independence assumption.} The primary conclusion of Section~\ref{sec:compounding}
is that validity failures compound multiplicatively. This conclusion depends on the
independence assumption: that layer failures are statistically unrelated. When the
same LLM provider family operates at all three layers, shared systematic biases
cause the cross-pipeline estimates of $V_i$ to overstate true validity for that
specific configuration, making the independence bound optimistic for same-provider
pipelines (\ref{app:compounding}, \S{}B.4). The qualitative conclusion
(failures multiply rather than add) is robust to this concern because the bound
holds as an upper limit regardless of provider overlap; same-provider configurations
are subject to an additional source of optimism that practitioners should account for.

\textit{Derived estimates in Table~\ref{tab:vtotal}.} The $V_1$ and $V_2$ values
in the benchmark table (Section~\ref{sec:worked-example}) are derived from
published measurements using approximations documented in the construct
validity subsection above; $V_3$ is not reported there because it is not a
well-defined quantity for a deterministic single-rule grader with no second
rater (\S\ref{sec:worked-example}).

Table~\ref{tab:sensitivity} below is therefore a hypothetical exercise, not an
extension of Table~\ref{tab:vtotal}'s measurements: it asks how $\vtot$ would
change \emph{if} a reliability-style validity coefficient of a given magnitude
could be assigned to the judgment layer, using $\tau$-bench and SWE-bench's
measured $V_1$/$V_2$ values as fixed inputs. It illustrates the sensitivity of
the compounding bound to the judgment layer, not a claim that $\kappa = 0.049$
\citep{gurram2026}, drawn from a different benchmark's substring-matching
grader, generalizes to either benchmark's actual (deterministic) grading
mechanism.

\begin{table}[htbp]
\centering
\caption{Hypothetical sensitivity of $\vtot$ to an assumed judgment-layer
coefficient for two representative benchmarks. $V_1$ and $V_2$ are held fixed
at the values in Table~\ref{tab:vtotal}; the three $V_3$ values are assumed
levels (chance-level, low, moderate), not measurements of either benchmark's
grader, whose $V_3$ is undefined under this model.}
\label{tab:sensitivity}
\small
\begin{tabular}{@{}lcccc@{}}
\toprule
Benchmark & $V_1$ & $V_2$ & $V_3$ & $\vtot \leq$ \\
\midrule
$\tau$-bench retail (SAE) & 0.62 & 0.88 & 0.05 & \textbf{0.027} \\
$\tau$-bench retail (SAE) & 0.62 & 0.88 & 0.30 & \textbf{0.164} \\
$\tau$-bench retail (SAE) & 0.62 & 0.88 & 0.50 & \textbf{0.273} \\
\midrule
SWE-bench & 0.59 & 1.00 & 0.05 & \textbf{0.030} \\
SWE-bench & 0.59 & 1.00 & 0.30 & \textbf{0.177} \\
SWE-bench & 0.59 & 1.00 & 0.50 & \textbf{0.295} \\
\bottomrule
\end{tabular}
\end{table}

The exercise shows how strongly the bound is governed by the judgment layer:
with $V_1$ and $V_2$ at their measured values, any assumed $V_3 \leq 0.50$
places $\vtot$ below the pipeline-redesign threshold proposed in
Section~\ref{sec:applying}, and only a judgment-layer coefficient above roughly
0.55 ($\tau$-bench) or 0.51 (SWE-bench) would lift the bound above 0.30. This is
a statement about the arithmetic of the bound under assumed inputs, not a claim
about either benchmark's actual validity, which cannot be computed until a
grader-vs-human validation study exists for its specific grader
(Section~\ref{sec:worked-example}).

% -----------------------------------------------------------------------
\section{Discussion}
\label{sec:discussion}

\subsection{What This Paper Does Not Claim}
\label{sec:not-claim}

We do not claim that current agentic evaluations are worthless. They provide
directional signal. We claim that the field's current practices overstate the
validity and precision of that signal, and that the standards for reporting should be
substantially higher. A benchmark result reported without IRR metrics, simulation
calibration evidence, and metric selection justification is incomplete, in the same
way that a clinical measurement reported without instrument validity data is
incomplete.

\subsection{The Governance Stakes}
\label{sec:governance}

For academic benchmark comparisons, validity limitations are a scientific quality
issue. For deployment decisions, safety certifications, and regulatory claims, they
are a governance failure. Agentic AI systems are increasingly being deployed in
consequential contexts. The evaluation apparatus used to justify those deployments
must meet the same validity standards applied to any measurement-driven governance
process.

The regulatory dimension makes this concrete. Emerging frameworks such as the EU~AI
Act require evidence of comprehensive risk assessment for general-purpose AI models.
A systematic analysis of 194,955 benchmark questions found that capabilities central
to loss-of-control (including evading human oversight, self-replication, and
autonomous AI development) receive zero benchmark coverage \citep{prandi2025}. If
the construct coverage gap means that no existing benchmark can provide evidence of
regulatory compliance, then the IRR validity of those benchmarks is a secondary
concern: the primary problem is that they do not measure what regulations require.
Both failures must be addressed together.

\subsection{Broader Impact}
\label{sec:broader-impact}

This work raises the evidential bar for deployment certifications, surfaces
demographic calibration failures that existing evaluation practice obscures, and
gives practitioners eight actionable criteria they can apply using existing
psychometric tools.

Three misuse risks follow from this work. First, the proposed reliability thresholds
($\alpha \geq 0.70$, $\alpha \geq 0.80$) may become compliance checkboxes rather
than genuine validity indicators. Meeting a threshold does not guarantee valid
measurement; it establishes the minimum condition for a score to be interpretable.
Second, stratified calibration across demographic and linguistic groups increases
evaluation cost; benchmark maintainers and shared evaluation platforms should bear
primary responsibility for calibration coverage, not individual research teams.
Third, the finding that three certified psychiatrists produced $\text{ICC} =
0.087$--$0.295$ on LLM mental health safety evaluation \citep{jafari2026} should not
discourage research in high-stakes AI domains. It means the construct is
underspecified for reliable measurement, which calls for more rigorous construct
development, not withdrawal.

% -----------------------------------------------------------------------
\section{Conclusion}
\label{sec:conclusion}

The move from static to dynamic agentic evaluation does not reduce the importance of
inter-rater reliability; it multiplies the surfaces on which reliability can fail
and introduces new systematic biases at each layer: task generation, world
simulation, and judgment. The compounding effect means that an evaluation pipeline
with moderate validity problems at each of three layers can produce an evaluation
signal that is far less valid than any single-layer estimate would suggest.

The science of measurement was not built for static benchmarks alone. It was built
for situations where the measurement apparatus is a source of variance, where raters
disagree in structured rather than random ways, and where the construct must be
separated from the instrument measuring it. That is agentic evaluation.

These standards are achievable. \citet{elhattami2025} (\textsc{WebArena Verified})
demonstrates this directly: adding rigorous human annotation with fixed annotator
pairs per site and reporting Cohen's $\kappa = 0.83$ (95\% CI [0.81, 0.85]) satisfies the
metric selection requirement (Prescription~2), clears the deployment-gating
reliability level (Prescription~5, read as the approximation noted in
Section~\ref{sec:decision-tree}), and treats IRR as a required reporting
field (Prescription~8). This is a demonstration of rigorous IRR \emph{practice}
for the construct it annotates (task-specification ambiguity and category
validity), not a benchmark-specific measurement of agent-outcome judgment
validity (Section~\ref{sec:worked-example}); the two are separable
achievements, and the field currently manages neither for most benchmarks.
\textsc{WebArena Verified} meets three of the eight prescriptions through a design
choice the field already knows how to make.

\fk, \ka, ICC, and construct validity frameworks are available, validated, and
applicable. The field's task is to use them, to report IRR as a first-class metric
alongside every benchmark result, and to recognize that a score without validity
evidence is not a measurement. It is a number.

% -----------------------------------------------------------------------
\section*{Acknowledgements}

The author thanks the agentic AI evaluation community whose published work,
including the benchmark papers, LLM-as-a-Judge studies, and calibration analyses
cited throughout, made this synthesis possible.

% -----------------------------------------------------------------------
% Elsevier mandatory declarations
% -----------------------------------------------------------------------

\section*{CRediT Authorship Contribution Statement}

\textbf{William Caban}: Conceptualization, Methodology, Formal Analysis,
Investigation, Writing -- Original Draft, Writing -- Review \& Editing,
Visualization.

\section*{Declaration of Competing Interests}

The author is employed by Red Hat, Inc. This research was conducted independently
as part of the author's doctoral program at Alma Mater Europaea University. The views
expressed are the author's own and do not represent the official position of Red Hat,
Inc. The author declares no financial competing interests.

\section*{Declaration of Generative AI and AI-Assisted Technologies}

During preparation of this work the author used Claude (Anthropic) to assist with
LaTeX formatting, editorial length reduction, and improvements of the
reliability-study code and analysis scripts; the author designed the study,
verified all outputs, and takes full responsibility for the work. Claude is not
listed as an author.

\section*{Funding}

No funding was received for this research.

\section*{Ethics Statement}

This study is based entirely on analysis of publicly available published research.
No human participants, personal data, or animal subjects were involved.

% -----------------------------------------------------------------------
\section*{Data Availability}

The data and code supporting this study are openly available at the repository
below. It contains the per-paper coding table for all 55 papers in the
literature scan (\texttt{full\_scan\_55papers.csv}, reproduced as
Table~\ref{tab:scan-full}) with the rationale recorded for every coded paper,
a machine-checkable evidence trail for the metric-reported code
(\texttt{audit\_report.csv}, which re-derives that code lexically from each
full document and reports where in the document the evidence sits), the author
codes and all five rater runs of the reliability study
(\texttt{author\_codes\_55.csv}, \texttt{results\_fulltext.csv}), and the Python scripts for extracting the
stimulus, running the LLM raters, validating the returned codes, and computing
Krippendorff's~$\alpha$ (\texttt{extract.py}, \texttt{second\_rater.py},
\texttt{validate.py}, \texttt{compute\_alpha.py}). Because the coded papers are
third-party copyrighted works, the repository publishes the SHA-256 checksum of
each extracted document (\texttt{stimulus\_hashes.csv}) and the pinned extractor
version rather than the extracted text itself, so that a replicator who obtains
the same PDFs can verify byte-identical stimulus reconstruction:

\begin{center}
\url{https://github.com/williamcaban/experiment-measurement-without-validity}
\end{center}

The repository is released under the Apache~2.0 license. All 55 coded papers are
publicly available; \texttt{manifest.csv} records the resolution scheme,
versioned identifier, fetch URL, and source-PDF checksum for each, which is what
makes the checksum verification above reproducible.
No proprietary datasets, model weights, or benchmark systems were created or used.
Code for the compounding model (\ref{app:compounding}) requires no
implementation beyond the equations stated.

% -----------------------------------------------------------------------
\bibliographystyle{elsarticle-num-names}
\bibliography{paper}

% -----------------------------------------------------------------------
\appendix
\setcounter{table}{0}
\renewcommand{\thetable}{A.\arabic{table}}

\section{Literature Scan: Coding Criteria, Category Summary, and Per-Paper Codes}
\label{app:literature}

This appendix documents the methodology and results of the structured literature scan
reported in Section~\ref{sec:scan}. It presents the pre-specified coding criteria, a
category summary of the 55 coded papers, the complete per-paper coding table, and
notable individual cases referenced in the main text.

\subsection*{A.1 Pre-Specified Coding Criteria}

Each of the 55 coded papers was coded on four dimensions, applied uniformly before reading
the paper's results section. Two codes were revised after the initial pass, both
on evidence from the mechanical lexical audit released as
\texttt{audit\_report.csv} rather than on reflection: one paper was found to
report a coefficient for its own annotation study that the first pass had
recorded as absent, and one was moved from absent to incomplete. The
\texttt{notes} column of the released coding table records the revision and its
date for each.

The dimensions are:

\begin{enumerate}
  \item \textbf{IRR metric reported}: Cohen's $\kappa$, Fleiss' $\kappa$,
    Krippendorff's $\alpha$, ICC (form specified or unspecified), percentage agreement
    only, or no metric reported.
  \item \textbf{Rater design}: number of raters; whether rater identity was fixed
    across items or varied (rotating pool, crowdsourced, or multiple independent runs
    of the same model).
  \item \textbf{Measurement scale}: binary, nominal (unordered categories), ordinal
    or ternary (ordered but discrete), or continuous.
  \item \textbf{Structural validity}: whether the reported metric's structural
    assumptions were satisfied by the rater design and measurement scale. Coding
    applied the following rules: \ck{} for $> 2$ raters or rotating identity $\to$
    \textsc{mismatch}; \fk{} for ordinal or continuous scale $\to$ \textsc{partial
    mismatch}; percentage agreement without chance correction $\to$
    \textsc{incomplete}; no IRR with automated grader of unvalidated accuracy $\to$
    \textsc{layer-3 validity failure}.
\end{enumerate}

\subsection*{A.2 Category Summary}

Table~\ref{tab:scan-summary} summarizes the 55 individually coded papers by topic
category. The nine categories were defined before coding to cover the main
structural situations in which IRR is, or should be, reported in agentic
evaluation: benchmarks scored automatically (A), studies that validate such
graders against humans (B), benchmarks that do report formal IRR (C),
LLM-as-a-Judge studies (D), benchmarks whose reported metric does not fit their
rater design (E), long-horizon and capability benchmarks (F), safety, RLHF, and
preference evaluation (G), recent 2025--2026 evaluation papers (H), and
safety-critical settings in which IRR failures carry direct consequences (I).
Categories A, C, and E were defined by an expected IRR pattern and therefore
contribute to the within-sample rate by design (Section~\ref{sec:internal-threats}).

\begin{table}[htbp]
\centering
\caption{Literature scan category summary. ``Correct'' = structurally valid metric
with explicit rationale. ``Failure mode'' = dominant pattern among incorrect papers.}
\label{tab:scan-summary}
\small
\begin{tabular}{@{}c>{\raggedright}p{0.33\linewidth}rr>{\raggedright\arraybackslash}p{0.27\linewidth}@{}}
\toprule
Cat. & Topic & Papers & Correct & Dominant failure mode \\
\midrule
A & Major agentic benchmarks (automated grading) & 9 & 0 & No IRR; automated grader validity assumed \\
B & Automated grader validity studies            & 3 & 1 & No IRR metric reported \\
C & Benchmarks with formal IRR                   & 11 & 9 & \ka{} applied to a self-consistency design; unweighted \ck{} on an ordinal scale \\
D & LLM-as-a-Judge studies                       & 4 & 0 & \ck{} aggregated over a 54-model panel; IRR metrics applied to self-consistency designs \\
E & Benchmarks with structural metric mismatch   & 5 & 0 & \fk{} for ordinal rubrics; \% only \\
F & Long-horizon and capability benchmarks       & 5 & 0 & No metric reported; \% agreement only \\
G & Safety, RLHF, and preference evaluation      & 8 & 0 & No IRR metric (ELO, Pearson, or none); \% agreement only \\
H & Recent 2025--2026 evaluation papers          & 3 & 0 & No metric; weighted \ck{} still not the prescribed metric \\
I & Safety-critical IRR failures                 & 7 & 1$^*$ & Raw \% agreement; coefficient reported without a stated threshold \\
\midrule
\textbf{Total} & & \textbf{55} & \textbf{11 (20\%)} & \\
\bottomrule
\multicolumn{5}{@{}l@{}}{\footnotesize $^*$ Category I includes one paper (\citealt{jafari2026}) that uses the correct metrics}\\
\multicolumn{5}{@{}l@{}}{\footnotesize but reports catastrophically low values (ICC $= 0.087$--$0.295$, $\alpha = -0.203$).}
\end{tabular}
\end{table}

\FloatBarrier
\subsection*{A.3 Per-Paper Coding Table}

Table~\ref{tab:scan-full} lists every paper in the scan with its four codes. The
year shown is the year of the version that was coded, which
\texttt{manifest.csv} pins to an exact identifier and checksum; where a work was
subsequently published, the reference list cites that version of record instead,
so the two can differ by a year for the same paper.
Q1 (metric reported): KCO = Cohen's $\kappa$; KFL = Fleiss' $\kappa$; KAL =
Krippendorff's $\alpha$; ICC = intraclass correlation; PA = percent agreement
only; NIL = no recognized IRR metric. Q2 (rater design): F2 = two fixed raters,
same pair on every item; FN = fixed panel of three or more; RV = rotating or
varying rater identity; CS = crowdsourced or open voting pool; UNK = not
described or no rater design. Q3 (scale): BIN, NOM, ORD, CON, UNK. Q4
(structural validity): OK = metric fits design and scale with threshold and
rationale; MM = structural mismatch; INC = percent agreement only, or correct
family without threshold or rationale; ABS = no IRR metric reported. The
rationale for each coded paper is recorded in the \texttt{notes} column of the
released coding table, and \texttt{audit\_report.csv} independently re-derives
the metric-reported code from each full document so that code can be checked
mechanically against its source.

\scriptsize
\begin{longtable}{@{}r>{\raggedright\arraybackslash}p{0.56\linewidth}cllll@{}}
\caption{Per-paper coding of the 55-paper literature scan. Category letters
follow Table~\ref{tab:scan-summary}.}
\label{tab:scan-full}\\
\toprule
\# & Paper & Cat. & Q1 & Q2 & Q3 & Q4 \\
\midrule
\endfirsthead
\multicolumn{7}{@{}l}{\textit{Table~\thetable\ (continued)}}\\
\toprule
\# & Paper & Cat. & Q1 & Q2 & Q3 & Q4 \\
\midrule
\endhead
\midrule
\multicolumn{7}{r@{}}{\textit{continued on next page}}\\
\endfoot
\bottomrule
\endlastfoot
1 & Yao et al. (2024), $\tau$-bench \citep{yao2024} & A & NIL & UNK & BIN & ABS \\
2 & Xie et al. (2024), OSWorld \citep{xie2024} & A & NIL & UNK & BIN & ABS \\
3 & Gurram (2026), AgentProp-Bench \citep{gurram2026} & B & KCO & F2 & BIN & OK \\
4 & El Hattami et al. (2025), WebArena Verified \citep{elhattami2025} & C & KCO & F2 & BIN & OK \\
5 & Megahed et al. (2025), Reliable decision support with LLMs & C & KAL & FN & BIN & OK \\
6 & Yamauchi et al. (2025), Empirical study of LLM-as-a-Judge design choices & C & KAL & UNK & ORD & MM \\
7 & Han et al. (2025), Judge's Verdict \citep{han2025} & D & KCO & RV & BIN & MM \\
8 & Jain et al. (2026), Multi-LLM thematic analysis with dual reliability metrics & D & KCO & UNK & NOM & MM \\
9 & Fan et al. (2026), AgentProcessBench \citep{fan2026} & E & KCO & F2 & ORD & MM \\
10 & Akkil et al. (2026), Emergence WebVoyager & E & PA & F2 & BIN & INC \\
11 & Hu, T. et al. (2025), Multi-agent debate for LLM judges & F & NIL & FN & CON & ABS \\
12 & Zhou, H. et al. (2026), Taxonomic bias evaluation of LLM judges & F & NIL & RV & ORD & ABS \\
13 & Movva et al. (2024), Annotation alignment (conversational safety) \citep{movva2024} & G & NIL & CS & ORD & ABS \\
14 & Chiang et al. (2024), Chatbot Arena \citep{chiang2024} & G & NIL & CS & BIN & ABS \\
15 & Chandwani et al. (2026), LH-Bench & H & KCO & FN & ORD & MM \\
16 & Ouyang et al. (2022), InstructGPT \citep{ouyang2022} & I & PA & CS & ORD & INC \\
17 & Jafari et al. (2026), Expert evaluation in mental-health AI safety testing \citep{jafari2026} & I & ICC & FN & ORD & OK \\
18 & Piot et al. (2025), LLM reliability in hate-speech detection (xRR) & I & KCO & RV & BIN & MM \\
19 & Kunilovskaya et al. (2026), Who annotates in NLP? & I & KAL & RV & NOM & INC \\
20 & James (2026), Counting on consensus (IAA metric selection) \citep{james2026} & I & NIL & UNK & UNK & ABS \\
21 & Jimenez et al. (2024), SWE-bench \citep{jimenez2024} & A & NIL & UNK & BIN & ABS \\
22 & Yang et al. (2026), HarmMetric Eval & B & NIL & UNK & BIN & ABS \\
23 & Ganguli et al. (2022), Red teaming language models to reduce harms \citep{ganguli2022} & E & KFL & FN & ORD & MM \\
24 & Seshadri et al. (2026), Lost in simulation \citep{seshadri2026} & F & NIL & UNK & UNK & ABS \\
25 & Meimandi et al. (2025), Measurement imbalance in agentic AI evaluation \citep{meimandi2025} & G & PA & F2 & BIN & INC \\
26 & Zhu et al. (2025), Best practices for rigorous agentic benchmarks \citep{zhu2025} & H & NIL & UNK & UNK & ABS \\
27 & Mialon et al. (2023), GAIA \citep{mialon2024} & A & NIL & UNK & UNK & ABS \\
28 & Zhou, S. et al. (2023), WebArena \citep{zhou2024webarena} & A & NIL & UNK & UNK & ABS \\
29 & Barres et al. (2025), $\tau^2$-Bench \citep{barres2025} & A & NIL & F2 & UNK & ABS \\
30 & Faisal et al. (2024), Dialectal toxicity detection \citep{faisal2024} & I & PA & UNK & ORD & INC \\
31 & Spiliopoulou et al. (2025), Play Favorites (self-bias in LLM-as-a-Judge) \citep{spiliopoulou2025} & C & KAL & FN & ORD & OK \\
32 & Wataoka et al. (2024), Self-preference bias in LLM-as-a-Judge \citep{wataoka2024} & D & NIL & UNK & UNK & ABS \\
33 & Bhardwaj et al. (2024), State of data curation at NeurIPS & C & ICC & RV & ORD & OK \\
34 & Pradhan et al. (2025), LLM-as-a-Judge for legal document recommendation & C & KAL & F2 & ORD & OK \\
35 & Marino et al. (2025), AIReg-Bench & C & KAL & RV & ORD & OK \\
36 & Guo et al. (2025), MCP-AgentBench & C & KFL & FN & BIN & OK \\
37 & Mazaheri et al. (2026), AgentAtlas & I & NIL & UNK & UNK & ABS \\
38 & Borse et al. (2025), IRR between LLMs and human raters in qualitative analysis & C & KCO & F2 & BIN & OK \\
39 & Xue et al. (2025), An illusion of progress? (web agents) & F & PA & F2 & BIN & INC \\
40 & Liu, P. et al. (2026), RLHF: a statistical perspective & G & NIL & UNK & UNK & ABS \\
41 & Zhao, W. et al. (2024), Challenges in trustworthy human evaluation of chatbots & E & KFL & FN & ORD & MM \\
42 & Coyne (2026), Three models of RLHF annotation & G & NIL & UNK & UNK & ABS \\
43 & Hu, Z. et al. (2024), Explaining length bias in LLM-based preference evaluations & G & KFL & RV & BIN & INC \\
44 & Frick et al. (2024), How to evaluate reward models for RLHF & G & NIL & CS & BIN & ABS \\
45 & Liu, S. et al. (2025), How humans help LLMs (preference annotators) & G & NIL & UNK & UNK & ABS \\
46 & Sharma et al. (2025), Consistent detection of cognitive distortions & E & KFL & RV & NOM & MM \\
47 & Keith (2025), LLM-as-a-Judge proxies for mathematical coherence & D & ICC & UNK & CON & MM \\
48 & Wu et al. (2026), Agent-based software artifact evaluation & C & KCO & F2 & ORD & MM \\
49 & Kwa et al. (2026), Measuring AI ability to complete long software tasks (METR) & A & NIL & UNK & UNK & ABS \\
50 & Lee et al. (2026), Meta-Harness & F & NIL & UNK & UNK & ABS \\
51 & Rabinovich et al. (2026), Near-Miss (latent policy failure detection) & B & NIL & UNK & BIN & ABS \\
52 & Yehudai et al. (2026), Survey on evaluation of LLM-based agents & H & NIL & UNK & UNK & ABS \\
53 & Libon et al. (2026), ResearchArena & A & NIL & UNK & UNK & ABS \\
54 & Sun et al. (2026), AgenticDataBench & A & NIL & UNK & UNK & ABS \\
55 & Wiyono et al. (2025), IndoPref & C & KAL & F2 & BIN & OK \\
\end{longtable}
\normalsize

\subsection*{A.4 Notable Individual Cases}

\begin{itemize}
  \item \citet{jafari2026}: Three certified psychiatrists, correct metrics, ICC
    $= 0.087$--$0.295$ and $\alpha = -0.203$ on LLM mental health response
    safety: highest-stakes domain, most disagreement.
  \item \citet{ouyang2022}: 40 rotating contractors, 4-way preference ranking.
    Reports 72--77\% raw percent agreement with no chance correction; requires
    Krippendorff's $\alpha$ for rotating-rater ordinal design.
  \item \citet{gurram2026}: Direct grader-vs-human comparison. Substring
    grading achieves $\kappa = 0.049$ (chance level), a 3-LLM ensemble
    $\kappa = 0.432$, a single GPT-4o-mini judge $\kappa = 0.567$, against
    $\kappa = 0.835$ between the two human annotators.
  \item \citet{elhattami2025} (\textsc{WebArena Verified}): One primary annotator
    per site with a paired blind verifier (four annotators across the four
    self-hosted sites; Map is not self-hosted and is excluded from their
    reliability analysis),
    task-specification-ambiguity and category-validity labels, Cohen's
    $\kappa = 0.83$ (95\% CI [0.81, 0.85]) task-weighted across all 812 tasks: correct
    metric, threshold exceeded, reporting complete. The strongest positive
    example of IRR \emph{practice} in the scan, though the labeled construct is
    benchmark-audit quality, not agent-outcome judgment (Section~\ref{sec:worked-example}).
\end{itemize}

\subsection*{A.5 Production Rules for IRR Metric Selection}

Table~\ref{tab:production-rules} states the decision procedure of
Figure~\ref{fig:decision-tree} as a rule base. It is given here rather than
beside the figure so that the body carries the procedure and the appendix
carries its formal statement; the two are the same artifact, and a practitioner
needs only the figure.

\begin{table}[htbp]
\centering
\caption{Production rules corresponding to the terminal nodes of the IRR metric
selection decision tree (Figure~\ref{fig:decision-tree}). Rules are applied in
order; the first matching rule determines the metric. R3 and R4 apply whether
the panel is fixed or rotates, provided the number of raters per item is
constant (\S\ref{sec:irr-metrics}); R2 is reserved for designs where that count
itself is irregular.}
\label{tab:production-rules}
\small
\begin{tabular}{@{}p{0.06\linewidth}p{0.52\linewidth}p{0.30\linewidth}@{}}
\toprule
Rule & Condition & Metric \\
\midrule
R1 & IF rater\_count = 2 AND identity\_fixed = true AND complete\_matrix = true
   AND scale = nominal/binary
   & Cohen's $\kappa$ \\[4pt]
R1$^\dagger$ & IF rater\_count = 2 AND identity\_fixed = true AND complete\_matrix = true
   AND scale = ordinal
   & Weighted $\kappa$ or Krippendorff's $\alpha$ (ordinal) \\[4pt]
R1$^\ddagger$ & IF rater\_count = 2 AND identity\_fixed = true AND complete\_matrix = true
   AND scale $\in$ \{interval, continuous\}
   & $\text{ICC}(A,1)$ \\[4pt]
R2 & IF missing\_ratings = true OR raters\_per\_item\_varies = true
   & Krippendorff's $\alpha$ (general) \\[4pt]
R3 & IF rater\_count $\geq$ 3 AND complete\_matrix = true AND
   raters\_per\_item\_constant = true AND scale $\in$ \{nominal, binary\}
   & Fleiss' $\kappa$ \\[4pt]
R4 & IF rater\_count $\geq$ 3 AND complete\_matrix = true AND
   raters\_per\_item\_constant = true AND scale $\in$ \{ordinal, interval, continuous\}
   & Krippendorff's $\alpha$ (ordinal/interval distance) \\
\bottomrule
\end{tabular}
\end{table}

% -----------------------------------------------------------------------
\section{Formal Derivation of the Compounding Validity Model}
\label{app:compounding}

\textit{This appendix provides the mathematical treatment supporting
Section~\ref{sec:compounding}.}

\subsection*{B.1 Definitions}

Let an agentic evaluation pipeline $\Pi$ consist of three sequential components:
\begin{itemize}
  \item $G$: a task generation function that samples tasks $t$ from a distribution
    $D_G$ over task space $\mathcal{T}$
  \item $S$: a simulation function that, given a task $t$ and an agent $A$, generates
    an interaction trajectory $\tau = S(t, A)$ drawn from a distribution $D_S(t)$
  \item $J$: a judgment function that, given a trajectory $\tau$, assigns a score
    $s = J(\tau) \in \mathbb{R}$
\end{itemize}

Let $\mathcal{C}^*$ denote the target construct: the latent capability the
evaluation is designed to measure. The construct $\mathcal{C}^*$ induces an ideal
task distribution $D_{\mathcal{C}^*}$ over $\mathcal{T}$ and an ideal scoring
function $J^*$ that maps agent behavior to true capability.

\subsection*{B.2 Layer-Specific Validity Measures}

\paragraph{$\cone$: Task Generation Validity.}
\[
  \cone = 1 - d_{\mathrm{TV}}(D_G,\, D_{\mathcal{C}^*})
\]
where $d_{\mathrm{TV}}$ denotes total variation distance. $\cone = 1$ when
$D_G = D_{\mathcal{C}^*}$ exactly; $\cone = 0$ when the supports are disjoint.

\paragraph{$\ctwo$: Simulation Calibration Validity.}
\[
  \ctwo = \mathrm{ICC}(\text{outcomes}_{\mathrm{sim}},\;
                         \text{outcomes}_{\mathrm{real}})
\]
where ICC is computed using the two-way mixed-effects model for absolute agreement
($\mathrm{ICC}(A,1)$). $\ctwo$ must be estimated separately per demographic and
linguistic user group:
\[
  \ctwo = \mathbb{E}_{g \sim P_{\mathrm{users}}}[\ctwo^{(g)}]
\]

\paragraph{$\cthree$: Judgment Validity.}
$\cthree$ measures the degree to which the rating protocol produces valid
inter-rater agreement, using the appropriate IRR metric for the pipeline's
measurement scale.

\subsection*{B.3 The Multiplicative Validity Bound}

\textbf{Conceptual bound.} We state the following as a conceptual model, not a
proved mathematical proposition. Under the assumption that $\cone$, $\ctwo$,
$\cthree$ are independently determined:
\[
  \vtot(\Pi) \leq \cone \cdot \ctwo \cdot \cthree
\]

\textit{Motivating argument.} Let $q$ denote an agent query drawn from the evaluation
pipeline. Under independence, the signal-to-noise ratio of the pipeline degrades
multiplicatively across layers:
\[
  \mathrm{SNR}(\Pi) \leq \mathrm{SNR}(G) \cdot \mathrm{SNR}(S) \cdot \mathrm{SNR}(J)
\]
Treating each $V_i$ as the normalized SNR of the corresponding component then yields
the bound. The motivating argument is that layer-wise validity losses multiply; this
intuition is well-grounded even if the full formal derivation is not supplied here.

\begin{remark}
Because this is a conceptual bound rather than a proved proposition, equality in the
inequality is not established. The bound is tightest (most informative) when the
cross-pipeline $V_i$ estimates accurately reflect the specific pipeline being
assessed. When shared systematic biases are present (for example, when the same
provider family operates at all three layers), the cross-pipeline estimates are
optimistic, making the bound itself loose (see \S{}B.4).
\end{remark}

\subsection*{B.4 Estimator Bias in Same-Provider Pipelines}

The point estimates $\cone$, $\ctwo$, $\cthree$ used in practice (e.g.,
Table~\ref{tab:vtotal}) are drawn from studies conducted on pipelines that differ
from the one being assessed. These estimates are therefore cross-pipeline
measurements, not direct measurements of the specific pipeline under evaluation.

When the same provider family operates at all three layers of a specific pipeline,
systematic tendencies (verbosity preference, formatting bias, positional
sensitivity) manifest simultaneously at $G$, $S$, and $J$. Because the cross-pipeline
estimates were obtained on pipelines without this shared bias, they do not capture
the additional validity loss attributable to within-family amplification. The
result is that the external $V_i$ estimates overstate the true layer validity for
that specific configuration, making the bound $\vtot \leq \cone \cdot \ctwo \cdot
\cthree$ optimistic: the true $\vtot$ for a same-provider pipeline may fall below
the bound implied by the independently-sourced estimates.

This effect operates through \emph{estimator bias}, not through statistical
dependence among the $V_i$ as random variables. The bound in Equation~\eqref{eq:bound}
remains valid as an upper bound regardless of provider overlap; the point is that
same-provider configurations produce cross-pipeline estimates that are systematically
too high, so the bound, while mathematically correct, is a loose one.

\textbf{Mitigation.} Using distinct provider families for task generation ($G$),
simulation ($S$), and judgment ($J$) ensures that shared systematic biases are not
amplified across layers. Cross-provider pipeline designs also allow cross-family
agreement to be measured directly as a validity check on the $V_3$ estimate.

\subsection*{B.5 Worked Numerical Example}

The values below are illustrative, not measurements. $\cone$ and $\ctwo$ are
plausible figures of the kind reported in Table~\ref{tab:vtotal}, while the two
$\cthree$ values are \emph{assumed} levels: no benchmark in this survey
publishes a judgment-layer coefficient, and for a deterministic grader
$\cthree$ is undefined under this model
(Section~\ref{sec:worked-example}). The example shows how the bound propagates
under an assumption about $\cthree$, not what any pipeline's validity is.

\textbf{Cross-provider design:}
$\vtot \leq 0.70 \times 0.75 \times 0.75 = 0.394$

\textbf{Within-family design with metric mismatch:}
$\vtot \leq 0.70 \times 0.75 \times 0.58 = 0.305$

An evaluation score reported to two decimal places from a pipeline in the
0.30--0.40 validity range is precise to a degree of resolution the underlying
measurement cannot support.

\subsection*{B.6 Estimation Protocol}

The practical estimation protocol for $V_1$, $V_2$, $V_3$, and $\vtot$ is
presented in full as Section~\ref{sec:applying} in the main body. The steps in
that section implement the formal definitions in B.1--B.2 using tractable
approximations: expert-rated sampling for $V_1$, $\mathrm{ICC}(A,1)$ between
simulated and real user outcomes for $V_2$ (estimated separately per demographic
group), and the structurally correct IRR metric from Figure~\ref{fig:decision-tree}
for $V_3$. The interpretation guideline ($\vtot \leq 0.50$: no consequential
decisions without independent validation; $\vtot \leq 0.30$: pipeline redesign
required) accompanies the protocol in Section~\ref{sec:applying}.

\end{document}